\documentclass[10pt,journal,compsoc]{IEEEtran}

\ifCLASSINFOpdf
\else
\fi

\usepackage{psfrag,amsbsy,graphics,float}
\usepackage{verbatim} 
\usepackage{url}
\usepackage{cite}
\usepackage[numbers]{natbib}

\usepackage{wrapfig}

\usepackage{algorithm}
\usepackage[noend]{algorithmic}
\usepackage{amsmath}
\usepackage{amssymb}
\usepackage{amsthm}
\usepackage{bbm}
\usepackage{enumerate}
\usepackage{graphicx}
\usepackage{ifthen}
\usepackage{struktex}
\usepackage{hyperref}
\usepackage{color}
\usepackage{nicefrac} 
\usepackage{multirow}
\usepackage{hhline}
\usepackage{cleveref}
\usepackage{tabularx}
\usepackage{rotating}
\usepackage{caption}
\usepackage{subcaption}
\usepackage{rotating}
\usepackage{listings}
\usepackage{xcolor}
\usepackage{xtab}

\lstdefinestyle{Python}{
    language        = Python,
    basicstyle      = \ttfamily,
    keywordstyle    = \color{blue},
    keywordstyle    = [2] \color{teal}, %
    stringstyle     = \color{green},
    commentstyle    = \color{red}\ttfamily
}

\usepackage{booktabs}
\usepackage{makecell}
\usepackage{longtable}
\usepackage[nolist]{acronym}

\usepackage{adjustbox}

\usepackage{soul}
\usepackage[normalem]{ulem}

\usepackage{newfloat}
\DeclareFloatingEnvironment[name={Supplementary Figure},fileext=lsf,listname={List of Supplementary Figures}]{suppfigure}

\DeclareFloatingEnvironment[name={Supplementary Tables},fileext=lsf,listname={List of Supplementary Tables}]{supptable}

\newcommand{\ie}{i.\,e.}
\newcommand{\eg}{e.\,g.}

\newcommand{\ds}{\mbox{\textsc{Deep Spectrum }}}

\begin{document}
\title{Automatic Emotion Modelling in Written Stories}

\author{Lukas~Christ$^{1}$,
        Shahin~Amiriparian$^{1}$,
        Manuel Milling$^{1}$,
        Ilhan Aslan$^{2}$\\
        and~Bj\"orn~W.\ Schuller$^{1,3}$,~\IEEEmembership{Fellow,~IEEE}%
        \thanks{$^{1}$Lukas Christ, Shahin Amiriparian, Manuel Milling and Bj\"{o}rn Schuller are with the Chair of Embedded Intelligence for Health Care \& Wellbeing, University of Augsburg, Germany. 
{\tt\small \{lukas.christ, shahin.amiriparian\}@informatik.uni-augsburg.de
}}
\thanks{$^{2}$Ilhan Aslan is with the Device Software Lab, Huawei Technologies, Germany {\tt\small ilhan.aslan@huawei.com}}
\thanks{$^{3}$Bj\"{o}rn Schuller is also with GLAM -- the Group on Language, Audio \& Music, Imperial College London, UK.}}

\markboth{Transactions on Affective Computing,~Vol.~xx, No.~x, x~2022}{}%

\IEEEtitleabstractindextext{%

\begin{abstract}
    Telling stories is an integral part of human communication which can evoke emotions and influence the affective states of the audience. %
    Automatically modelling emotional trajectories in stories has thus attracted considerable scholarly interest. However, as most existing works have been limited to unsupervised dictionary-based approaches, there is no labelled benchmark for this task. We address this gap by introducing continuous valence and arousal annotations for an existing dataset of children's stories annotated with discrete emotion categories. We collect additional annotations for this data and map the originally categorical labels to the valence and arousal space. Leveraging recent advances in Natural Language Processing, we propose a set of novel Transformer-based methods for predicting valence and arousal signals over the course of written stories. We explore several strategies for fine-tuning a pretrained \textsc{ELECTRA} model and study the benefits of considering a sentence's context when inferring its emotionality. Moreover, we experiment with additional \ac{LSTM} and Transformer layers. The best configuration achieves a Concordance Correlation Coefficient (CCC) of $.7338$ for valence and $.6302$ for arousal on the test set, demonstrating the suitability of our proposed approach. 
Our code and additional annotations are made available at \href{https://github.com/lc0197/emotion_modelling_stories}{https://github.com/lc0197/emotion\_modelling\_stories}.
\end{abstract}

}

\maketitle

\IEEEdisplaynontitleabstractindextext

\IEEEpeerreviewmaketitle

\IEEEraisesectionheading{\section{Introduction}\label{sec:introduction}}

Humans have been characterised as ``storytelling animals''~\cite{gottschall2012storytelling}, meaning that stories -- in a broad sense -- are a key aspect of both individual and social life. Stories are central to literature, movies, and music, but also to our dreams and memories. They contribute to shaping individual~\cite{maclean2015identity} and collective identity~\cite{humle2014remembering}. In the wake of the 
``narrative turn'', storytelling has received widespread attention from various disciplines for many decades~\cite{polletta2011sociology}. Research on storytelling has been conducted, \eg, in the fields of psychology~\cite{sunderland2017using}, anthropology~\cite{boyd2010origin}, cognitive sciences~\cite{burke2015neuroaesthetics}, and history~\cite{palombini2017storytelling}.

The practice of telling stories is arguably as old as humans' ability to communicate~\cite{anderson2010storytelling}. A crucial aspect of stories is their emotionality, as stories typically evoke a range of different emotions in the listeners or readers, which also serves the purpose of keeping the audience interested~\cite{hogan2011affective}. 

In recent years, several efforts have been made to model emotionality in written stories computationally.

However, as shown in~\Cref{sec:rw}, these studies have often been constrained to dictionary-based methods~\cite{reagan2016emotional, somasundaran2020emotion} and thus simple, static word representations. 
In addition, existing work often models emotions in stories on the sentence level only~\cite{agrawal2012unsupervised, batbaatar2019semantic} without taking into account surrounding sentences, missing out on important contextual information. 
In this study, we address the aforementioned issues by employing a pretrained Transformer~\cite{vaswani2017attention} model to automatically predict emotionality in stories. 
Transformers in the tradition of \textsc{BERT}~\cite{devlin2018bert} have achieved new state-of-the-art results in various \ac{NLP} tasks such as sentiment analysis~\cite{jiang2019smart, yang2019xlnet}, humour recognition~\cite{weller2019humor, christ2022multimodal} and \ac{TER}~\cite{acheampong2021transformer} and are thus a promising method for the task at hand.

In combination with an emotional \ac{TTS} system~\cite{triantafyllopoulos2022overview}, our system could serve naturalistic human-machine interaction, educational, and entertainment purposes~\cite{lugrin2010exploring}. For example, stories could be automatically read to children~\cite{eisenreich2014tale} by voice assistants in a car 
or other 
environment. Moreover, writing support systems~\cite{alabdulkarim2021automatic} could benefit from our method, assisting authors in expressing emotions in their texts. Furthermore, the prediction of emotions in literary texts is of interest in the field of Digital Humanities~\cite{kim2018survey}, especially in Computational Narratology~\cite{mani2014computational, piper2021narrative}.

We conduct our experiments on the children's story dataset created by~\citet{alm2008affect}. 
Specifically, our contributions are the following. First, we extend the annotations provided by~\citet{alm2008affect} and map the originally discrete emotion labels to the valence and arousal~\cite{russell1980circumplex} space (cf.~\Cref{sec:data}). We then explore a variety of deep learning and especially Transformer-based methods for predicting valence and arousal in the stories provided in the dataset (cf.~\Cref{sec:exp-setup}). To the best of our knowledge, our work is the first to model emotional trajectories in children's stories over the course of complete stories, also referred to as \emph{emotional arcs}, using supervised machine learning. While previous studies have typically focused solely on emotion prediction for single sentences, we show that considering the context of a sentence in a story notably improves prediction accuracy (cf.~\Cref{sec:results}).

\section{Related Works}\label{sec:rw}

Various unsupervised, lexicon-based approaches to model emotional trajectories in narrative and literary texts have been proposed. With a lexicon-based method,~\citet{reagan2016emotional} identified six elementary sentiment-based \emph{emotional arcs} such as \emph{rags-to-riches} in a corpus of about $1\,300$ books. Similarly,~\citet{somasundaran2020emotion} computed such arcs for narratives written by students via several dictionaries. Utilising the NRC emotion dictionary~\cite{mohammad2013crowdsourcing},~\citet{kim2017prototypical} linked different typical emotional trajectories to different genres of literature. Employing the same dictionary,~\citet{mohammad2012once} calculated densities of emotionally connotated words in different genres of text, fairytales among them. An unsupervised method based on WordNet-Affect~\cite{strapparava2004wordnet} to assign discrete emotions to sentences in fairytales was proposed by~\citet{mac2010evaluation} and built upon by~\citet{zad2020systematic}. While the previously mentioned studies focus on emotionality on a sentence or paragraph level,~\citet{elsner2012character} plotted dictionary-based~\cite{wilson2005recognizing} emotional trajectories for individual characters in novels, indicating the degree of emotionality associated with the character.~\citet{yavuz2020analyses} addressed a similar task, namely modelling character-level emotions in dramatic plays via the NRC Emotion Lexicon~\cite{mohammad2013crowdsourcing}.

Moreover, a range of datasets of narratives annotated for emotionality exists. 
In a corpus of $100$ crowdsourced short stories,~\cite{mori2019narratives} provided annotations both for character emotions as well as for emotions evoked in readers.
The \ac{DENS}~\cite{liu2019dens} contains about $10\,000$ passages from modern as well as classic stories, labelled with $10$ discrete emotions. In the authors' experiments, fine-tuning \textsc{BERT} proved to be superior to more classic approaches such as \acp{RNN}. 
The \ac{REMAN} dataset~\cite{kim2018feels} comprises $1\,720$ text segments from about $200$ books. These passages are labelled regarding emotion, the emotion experiencer, the emotion's cause and its target.~\citet{kim2018feels} conducted experiments with biLSTMs and \acp{CRF} on \ac{REMAN}. In~\cite{kim2019frowning}, the authors proposed the task of modelling emotional relations between characters in stories, presenting a suitable corpus from fan-fiction stories and experiments using \acp{RNN}. Further,~\citet{kim2019analysis} extended the \ac{REMAN} dataset to study non-verbal expressions of emotions in the contained stories.

The \ac{SEND}~\cite{ong2019modeling} is a multimodal dataset containing $193$ video clips of subjects narrating personal emotional events. These recordings are labelled with valence values in a time-continuous manner. 
~\citet{wu2019attending} experimented with different models for predicting valence in~\ac{SEND} based on the transcripts alone, achieving their best result with a Transformer variant. 

Another multimodal dataset of narratives is the Ulm State-of-Mind in Speech (USoMS) corpus~\cite{rathner2018state, schuller2018interspeech} consisting of $100$ audiovisual recordings, including textual transcripts of personal narratives about emotional events. It is annotated with the subjects' self-reported valence and arousal before and after telling their stories.~\citet{stappen2019context} trained an attention-based model on the provided transcripts in order to predict these valence and arousal values, discretised into three classes.

The corpus of children's stories~\cite{alm2008affect} we are using for our experiments is labelled for eight discrete emotions (cf.~\Cref{sec:data}).~\citet{alm2005emotional} modelled emotional trajectories in a subset of this corpus, while in~\cite{alm2005emotions}, the authors conducted machine learning experiments with several handcrafted features such as sentence length and POS-Tags as well as Bag of Words. The corpus has frequently served as a benchmark for textual emotion recognition. However, scholars have so far limited their experiments to subsets of this dataset, selected based on high agreement among the annotators or certain emotion labels. Examples of such studies include an algorithm combining vector representations and syntactic dependencies by~\citet{agrawal2012unsupervised}, the rule-based approach proposed by~\citet{udochukwu2015rule}, and a combination of \ac{CNN} and \ac{LSTM} introduced by~\citet{batbaatar2019semantic}. 
None of these works, however, aimed at modelling complete stories.

\section{Data}
\label{sec:data}
We opt for the children's story dataset by~\citet{alm2008affect}, as it is reasonably large, comprising about $15\,000$ sentences, and contains full, yet brief stories, with the longest story consisting of $530$ sentences.
 Moreover, the data is labelled per sentence, allowing us to model emotional trajectories for stories. We extend the dataset by a third annotation, as described in~\Cref{ssec:annotation}, and modify the originally discrete annotation scheme by mapping it into the continuous valence/arousal space (cf.~\Cref{ssec:mapping}).

Originally, the dataset comprises $176$ stories from $3$ authors. More precisely, $80$ stories from the German \textit{Brothers Grimm}, $77$ stories by Danish author \textit{Hans-Christian Andersen}, and $19$ stories written by \textit{Beatrix Potter} are contained. Every sentence is annotated with the emotion experienced by the primary character (\textit{feeler}) in the respective sentence, and the overall mood of the sentence. For both label types, annotators had to select one out of eight discrete emotion labels, namely \textit{anger}, \textit{disgust}, \textit{fear}, \textit{happiness}, \textit{negative surprise}, \textit{neutral}, \textit{positive surprise}, and \textit{sadness}.  Each sentence received labels from two different trained annotators. 

    For a detailed description of the original data, the reader is referred to~\cite{alm2005emotional, alm2008affect}. Statistics on the subset of the data we utilise in our experiments are provided in~\Cref{ssec:annotation}.

    With the emotional \ac{TTS} application in mind, we limit our experiments to predicting the mood per sentence, as it refers to the sentence as such instead of one particular subject.

\subsection{Additional Annotations}\label{ssec:annotation}
In addition to the existing annotations, we collect a third mood label for every sentence. This allows us to create a continuous-valued gold standard (cf.~\Cref{ssec:mapping}) via the agreement-based \ac{EWE}~\cite{grimm2005evaluation} fusion method, for which at least three different ratings are required. Compared to the original dataset, however, we opt for a reduced labelling scheme, eliminating both \textit{positive surprise} and \textit{negative surprise} from the set of emotions. We follow the reasoning of~\citet{susanto2020hourglass} and~\citet{ortony2022all}, who argue that \textit{surprise} in itself is not \emph{valenced}, \ie, of negative or positive polarity, but can only be polarised in combination with polar emotions. In other words, we would conceptualise, \eg, \textit{negative suprise} as a negative emotion such as \textit{anger}, \textit{disgust}, or \textit{fear} coupled with the neutral emotion of \textit{surprise}. Since surprise is thus not considered a basic emotion, we do not include it in our annotation scheme.

We calculate the Krippendorff agreements, ignoring the two different labelling schemes. Krippendorff's alpha ($\alpha$) for the whole dataset considering all three annotators is $.385$, when calculated on the basis of single sentences. The mean $\alpha$ per story is $\mu_\alpha=.341$, with a standard deviation of $\sigma_\alpha=.126$, indicating that the level of agreement is highly dependent on the story.
We remove stories whose $\alpha$ is smaller than $\mu_\alpha - 2\sigma_\alpha$. 
A detailed listing of $\alpha$ values for the remaining data on both the sentence and the story level is provided in~\Cref{tab:agreement}.

     \begin{table}[h!]
     \centering
     \resizebox{1\columnwidth}{!}{
\begin{tabular}{llrrrr}

\toprule 
Annotators & Level & \multicolumn{1}{r}{Overall} & \multicolumn{1}{r}{Grimms} & \multicolumn{1}{r}{HCA} & \multicolumn{1}{r}{Potter} \\ 
\midrule 

\multirow{2}{*}{A1,A2} & sent. & .356 & .272 & .411 & .333 \\ 
 & story & .297 ($\pm$.174) & .245 ($\pm$.196) &  .350 ($\pm$.149) & .307 ($\pm$.073) \\
 \midrule 

 \multirow{2}{*}{A1,A3} & sent. & .420 & .370 & .447 & .433 \\ 
 & story & .376 ($\pm$.184) & .346 ($\pm$.212) &  .395 ($\pm$.158) & .428 ($\pm$.126) \\
 \midrule 

 \multirow{2}{*}{A2,A3} & sent. & .383 & .331 & .408 & .391 \\ 
 & story & .338 ($\pm$.176) & .296 ($\pm$.172) &  .376 ($\pm$.178) & .3614 ($\pm$.139) \\
 \midrule 

 \multirow{2}{*}{A1,A2,A3} & sent. & .387 & .325 & .422 & .390 \\ 
 & story & .343 ($\pm$.126) & .301 ($\pm$.133) &  .380 ($\pm$.118) & .370 ($\pm$.062) \\
 \midrule 

\end{tabular}} \caption{$\alpha$ values for all possible combinations of annotators. The values are given for the whole dataset (\textit{Overall}) and the individual authors (\textit{Grimms, HCA, Potter}). The \textit{sent.} rows report the alphas on the basis of sentence annotations, in \textit{story} rows, the means, as well as standard deviations of alpha values per story, can be found.} \label{tab:agreement} 
     \end{table}
     
 \Cref{tab:agreement} clearly illustrates that agreement is also author-dependent, \eg, for all combinations of annotators, the sentence-wise agreement for the \textit{Grimm} brothers is lower than for both other authors. Even though the label set available to annotator 3 was reduced by two labels, the pairwise alphas involving annotator 3 are frequently higher than the pairwise alphas of the two original annotators, \eg, regarding the overall sentence-level agreement.

The removal of low-agreement stories leaves us with our final data set of $169$ stories. Key details of the data are summarised in~\Cref{tab:data}.

\begin{table}[]
    \centering
    \begin{tabular}{lrrrr}
    \toprule
         \multicolumn{1}{c}{} & \multicolumn{1}{c}{Overall} & \multicolumn{1}{c}{Grimm} & \multicolumn{1}{c}{HCA} & \multicolumn{1}{c}{Potter} \\ \midrule 

        \multicolumn{5}{l}{\textit{Size}} \\
        \# sentences & 14\,884 & 5\,236 & 7\,712 & 1\,936 \\
         \# stories & 169 & 77 & 73 & 19 \\ 
\cmidrule{0-0}
\multicolumn{5}{l}{\textit{Emotion Labels (\%)}} \\
         anger & 4.54 & 6.71 & 2.77 & 5.71 \\

         disgust & 2.35 & 1.78 & 2.83 & 1.98 \\

         fear & 7.21 & 11.48 & 3.77 & 9.38 \\

         happiness & 14.42 & 13.74 & 16.59 & 7.58\\

         negative surprise & 4.41 & 4.17 & 4.74 & 3.72 \\

         neutral & 56.19 & 49.88 & 57.86 & 66.56 \\

         positive surprise & 1.90 & 2.73 & 1.54 & 1.08 \\

         sadness & 8.99 & 9.51 & 9.89 & 3.97 \\
         \bottomrule
    \end{tabular}
    \caption{Key statistics for the entire dataset and the subsets defined by the three different authors.}
    \label{tab:data}
    
\end{table}

The label distribution statistics listed in~\Cref{tab:data} point to stylistic differences between the different authors. To give an example, in the stories of \textit{Potter}, $66.56\,\%$ of all annotations are neutral, while for the \textit{Grimm} brothers only $49.88\,\%$ of them are. On the other hand, \textit{sadness} seems to be rare in \textit{Potter's} stories ($3.97\,\%$ of all annotations) compared to the other two authors. Both the overall and the individual authors' class distributions are highly skewed, with \textit{neutral} being the most frequent label, while other classes, especially \textit{positive surprise} and \textit{disgust}, are underrepresented.

\begin{figure}[h!]
    \centering
    \includegraphics[width=\columnwidth]{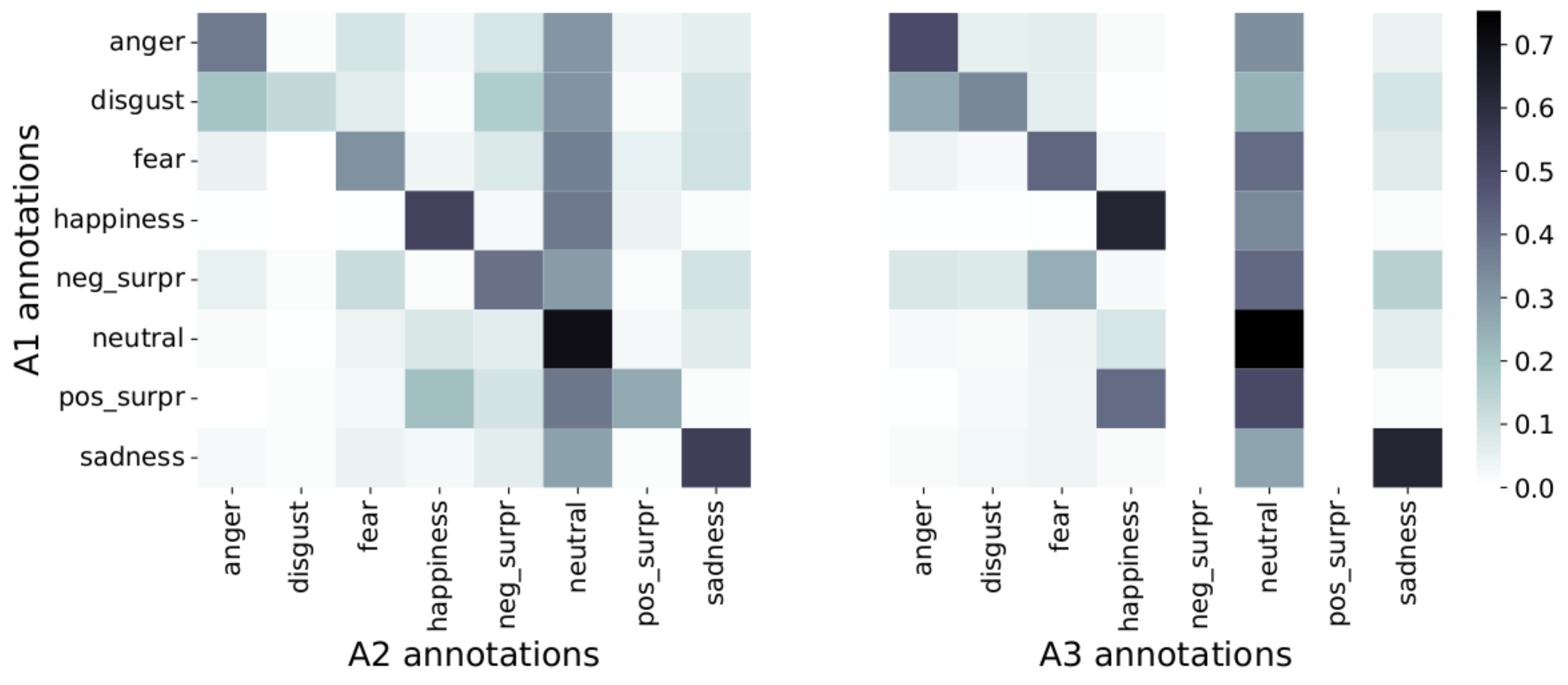}
    \caption{Confusion matrices comparing different annotators' (A1, A2, A3) labels for the whole dataset. Note that for annotator 3, \textit{positive} and \textit{negative surprise} were not available.}
    \label{fig:conf}
\end{figure}

\Cref{fig:conf} shows confusion matrices comparing the annotations of annotator 1 with the annotations of annotators 2 and 3.

Clearly, the decision whether a sentence is emotional or \textit{neutral} is the most important source of disagreement in both comparisons. Furthermore,~\Cref{fig:conf} demonstrates that disagreement about the valence of a sentence's mood is rare. To give an example, in both depicted confusion matrices, sentences labelled with \textit{happiness} by annotator 1 are almost never labelled with a negative emotion (\textit{anger}, \textit{disgust}, \textit{fear}) by annotator 2 and 3, respectively.

\subsection{Label Mapping}\label{ssec:mapping}
Motivated by low to moderate Krippendorff agreements (cf.~\Cref{tab:agreement}) and underrepresented classes in the discrete annotations (cf.~\Cref{tab:data}), we project all labels into the more generic, continuous valence/arousal space. Indeed, from~\Cref{fig:conf}, it is clear that annotators often agree on the polarity of the emotion. Hence, it can be argued that disagreement between annotators is not always as grave as suggested by the Krippendorff values in~\Cref{tab:agreement}, which do not take proximity between different emotions into account. To give an example, disagreement on whether a sentence's mood is \textit{happiness} or \textit{neutral} is certainly less severe than one annotator labelling the sentence \textit{sad}, while the other opts for \textit{happy}. Moreover, a projection into continuous space unifies the two different label spaces defined by the original and our additional annotations, respectively.

While supervised learning for predicting continuous valence and arousal signals is common in the field of multimodal affect analysis~\cite{ringeval2019avec, stappen2021muse, Christ22-TM2}, it has not been applied to textual stories, yet. Emotional arcs as discussed in~\Cref{sec:rw} are typically only valence-based and are computed without employing supervised learning techniques, with the exception of~\cite{wu2019attending}. 

To implement the desired mapping, we take up an idea proposed by~\citet{park-etal-2021-dimensional}, who map discrete emotion categories to valence and arousal values by looking up the label (\eg, \textit{anger}) in the NRC-VAD dictionary~\cite{mohammad2018obtaining}. However, the dictionary does not contain entries for \textit{positive surprise} and \textit{negative surprise}. For \textit{positive surprise}, we take the valence and arousal values of \textit{surprise} (both $.875$). The valence value for \textit{negative surprise} is set to the mean valence value of the negative emotions \textit{anger}, \textit{disgust},  and \textit{fear} ($.097$), while the arousal value is the same as for \textit{positive surprise} ($.875$).~\Cref{tab:mapping} lists the mapping for all discrete emotion labels.
    \begin{table}[h!]
\centering
\begin{tabular}{lrr}
\toprule 
\multicolumn{1}{l}{Label} &  \multicolumn{1}{r}{Valence} & \multicolumn{1}{r}{Arousal} \\ 

\midrule

Anger & .167 & .865 \\
Disgust & .052 & .775 \\
Fear & .073 & .840 \\
Happiness & .960 & .732 \\
Negative Surprise & .097 & .875 \\
Neutral & .469 & .184 \\
Positive Surprise & .875 & .875 \\
Sadness & .052 & .288 \\
\bottomrule
\end{tabular}
\caption{Mapping from discrete labels to continuous valence and arousal values.}\label{tab:mapping}
\end{table}
    
\begin{figure*}[h!]
    \centering
    \includegraphics[width=.85\linewidth, page=1]{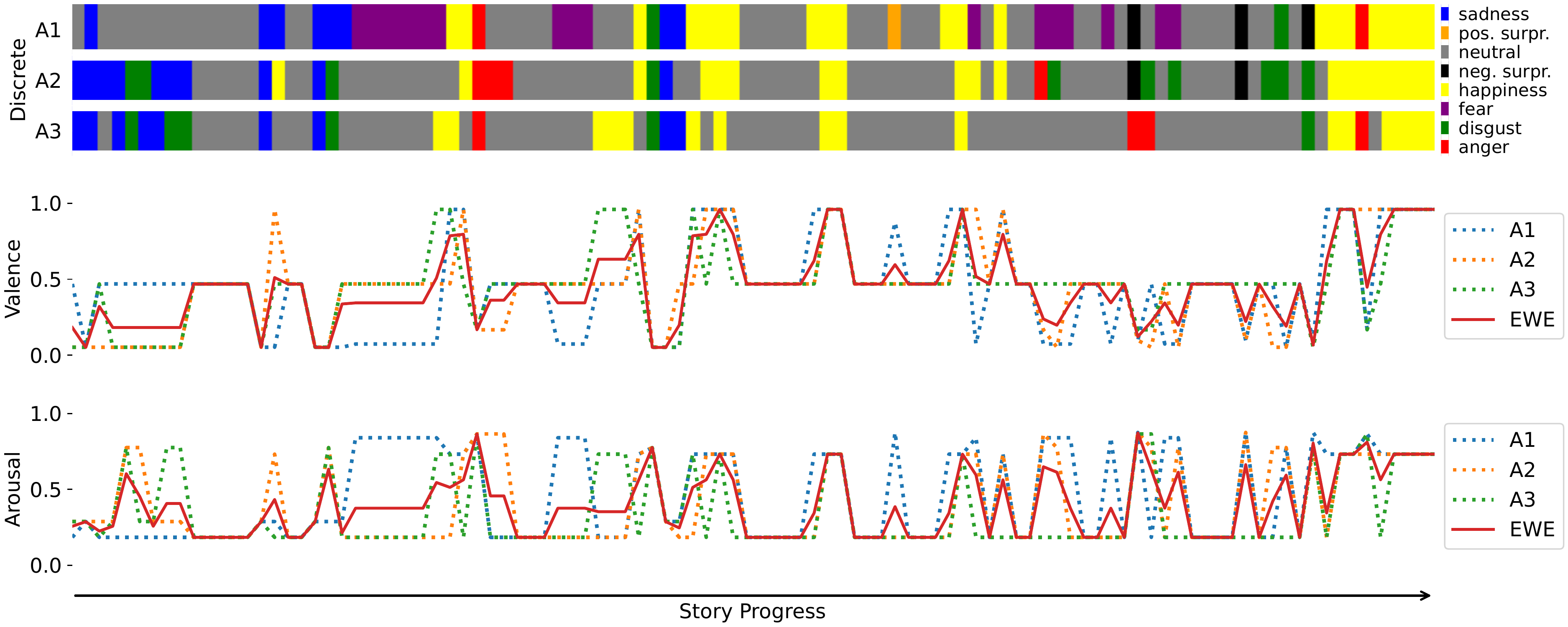}

    \caption{Exemplary mapping from the three annotators' (A1, A2, A3) discrete annotations (top) to their respective valence (middle) and arousal (bottom) signals and the gold standard signals created via EWE (solid red lines). The annotations are taken from the story \textit{Ashputtel} by the \textit{Grimm} brothers, consisting of 102 sentences.}\label{fig:signals}
\end{figure*}

    Having mapped the three labels assigned to each sentence into valence/arousal space according to~\Cref{tab:mapping}, we create a gold standard for every story by fusing the thus obtained signals over the course of a story for valence and arousal, respectively. We apply the \ac{EWE}~\cite{grimm2005evaluation} method which is well-established for the problem of computing valence and arousal gold standards from continuous signals (\eg,~\cite{ringeval2019avec, stappen2021muse, Christ22-TM2}).~\Cref{fig:signals} presents an example for this process, presenting both the discrete labels and the valence and arousal signals constructed from them for a specific story.

\subsection{Splits}\label{ssec:splits}
As the original dataset does not provide any data partitions, we create our splits on the level of stories.
In doing so, we make sure to include comparable portions of stories and sentences by each author in all three partitions. Our training partition comprises $118$ stories, and the development and test partitions contain $25$ and $26$ stories, respectively. A detailed breakdown is displayed in~\Cref{tab:partition}. We refer to this data split as the \textit{main split} in the remainder of this paper.

\begin{figure}[h!]
    \centering
    \subfloat[Valence Values]{
    \includegraphics[width=.47\columnwidth]{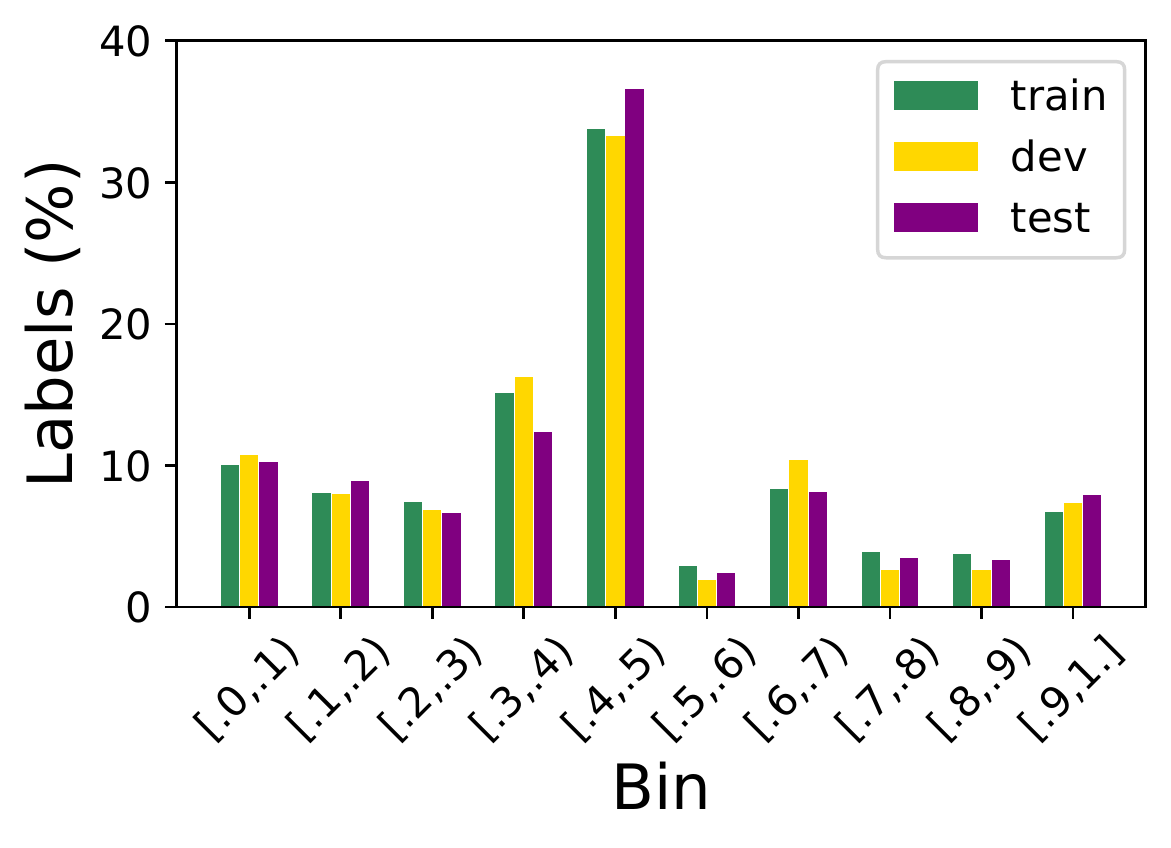}}
    \subfloat[Arousal Values]{
    \includegraphics[width=.47\columnwidth]{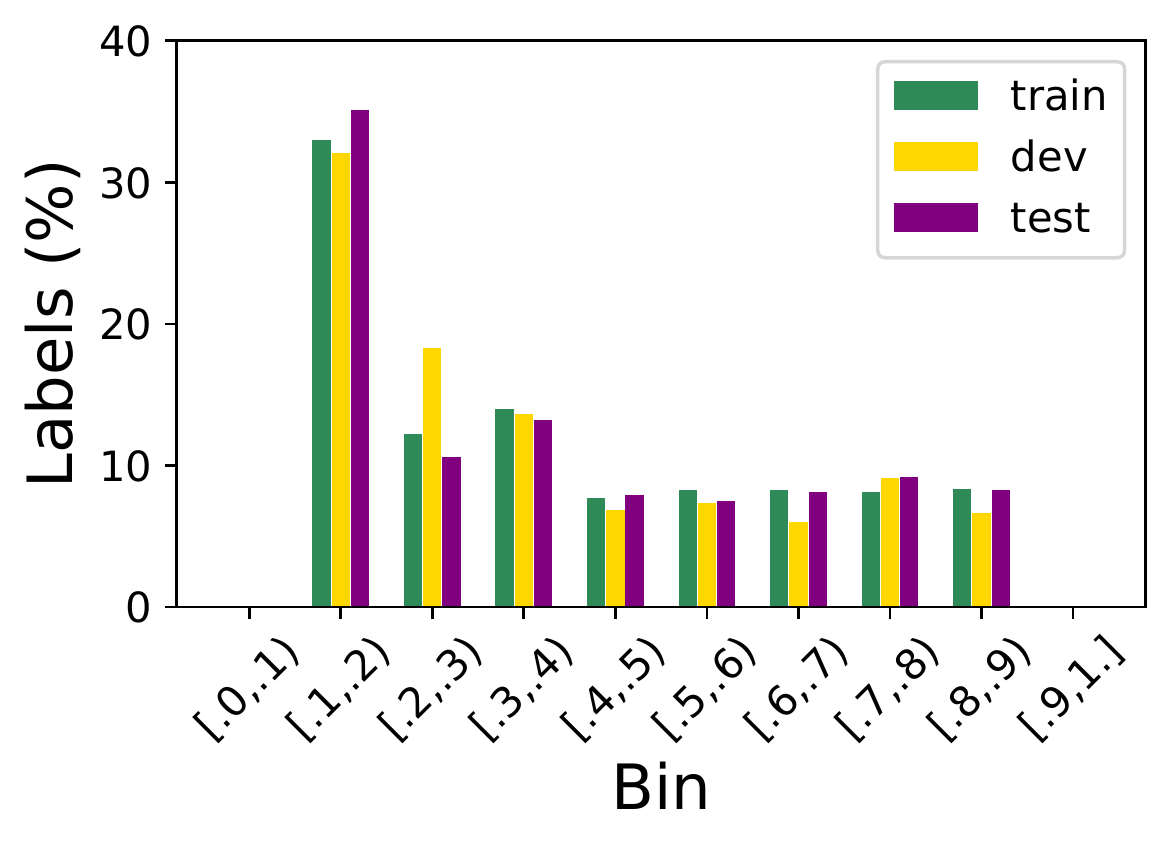}}
    \caption{Distributions of binned valence and arousal values in the created training, development (\textit{dev}) and test partition.}
    \label{fig:split_bins}  
\end{figure}

\begin{table}[h!]
    \centering\resizebox{1\columnwidth}{!}{
    \begin{tabular}{lrrrr}
    \toprule
         & \multicolumn{1}{c}{Overall} & \multicolumn{1}{c}{Grimm} & \multicolumn{1}{c}{HCA} & \multicolumn{1}{c}{Potter} \\ \midrule

         \multicolumn{5}{l}{\textbf{train}} \\ 
         
         stories & 118 & 54 (45.76\,\%) & 51 (43.22\,\%) & 13 (11.02\,\%) \\ 
         sentences & 10\,121 & 3\,621 (35.78\,\%) & 5\,246 (51.38\,\%) & 1\,254 (12.39\,\%) \\
         \midrule

\multicolumn{5}{l}{\textbf{development}} \\
         stories & 25 & 9 (36.00\,\%) & 13 (52.00\,\%) & 3 (12.00\,\%) \\
         sentences & 2\,384 & 604 (25.34\,\%) & 1\,494 (58.47\,\%) & 386 (16.19\,\%) \\
         \midrule
         
\multicolumn{5}{l}{\textbf{test}} \\
         stories & 26 & 14 (53.85\,\%) & 9 (34.62\,\%) & 3 (11.54\,\%) \\
         sentences & 2\,379 & 1\,011 (42.50\,\%) & 1\,072 (45.06\,\%) & 296  (12.44\,\%) \\
         \bottomrule
    \end{tabular}}
    \caption{Dataset split statistics for every partition and author. For each author, the absolute number of stories as well as sentences in each partition is given. The percentage values denote the share of the author's stories/sentences in the stories/sentences of the respective partition.}
    \label{tab:partition}
\end{table}
\Cref{fig:split_bins} shows that the continuous label distributions are fairly similar in the different partitions. 
Since all stories were written by only three different authors, the dataset also provides the opportunity to evaluate our models on the author level. We thus also experiment with author-based splits, where each of the three partitions corresponds to the stories of one author.

\section{Experimental Setup}
\label{sec:exp-setup}

Fine-tuning Transformer-based Language Models has become the standard method for various \ac{NLP} tasks such as sentiment analysis and predicting semantic similarity in recent years~\cite{devlin2018bert, lan2019albert, yang2019xlnet}. For the task at hand, we opt for \textsc{ELECTRA}~\cite{clark2020electra} which proved to be superior over the vanilla \textsc{BERT}~\cite{devlin2018bert} approach in several \ac{NLP} problems including sentiment analysis~\cite{clark2020electra} and discrete emotion recognition~\cite{kim-etal-2020-contextual}. We fine-tune \textsc{ELECTRA} models (cf.~\Cref{ssec:finetuning}) and combine them with \acp{RNN} and additional Transformer layers (cf.~\Cref{ssec:context_modelling}). In all experiments, we use the training set to train the respective models while the development set is employed to monitor the model's performance after each training epoch for the purpose of early stopping.

\subsection{Finetuning ELECTRA}\label{ssec:finetuning}
Since the context of a sentence in a story is often relevant to the mood it conveys, we leverage the surrounding of a sentence in the fine-tuning process. The input format for the pretrained \textsc{ELECTRA} model is defined as \verb|[CLS] text1 [SEP] text2|, where \verb|[CLS]| is a special token whose embedding is intended to represent the sentence \verb|text1| and additional text \verb|text2| can be provided after the special token \verb|[SEP]|. Accordingly, we feed our training data into the model as \verb|[CLS] sentence [SEP] context| for each sentence in the training partition. We add a feed-forward layer on top of \textsc{ELECTRA}'s representation of the \verb|[CLS]| token. It projects the $768$-dimensional embedding to $2$ dimensions and is followed by Sigmoid activation for both of them, corresponding to a prediction for valence and arousal of \verb|sentence|, respectively. As the loss function, we sum up the \acp{MSE} for valence and arousal. 

\begin{figure}[h!]
    \centering
    \includegraphics[width=.95\columnwidth]{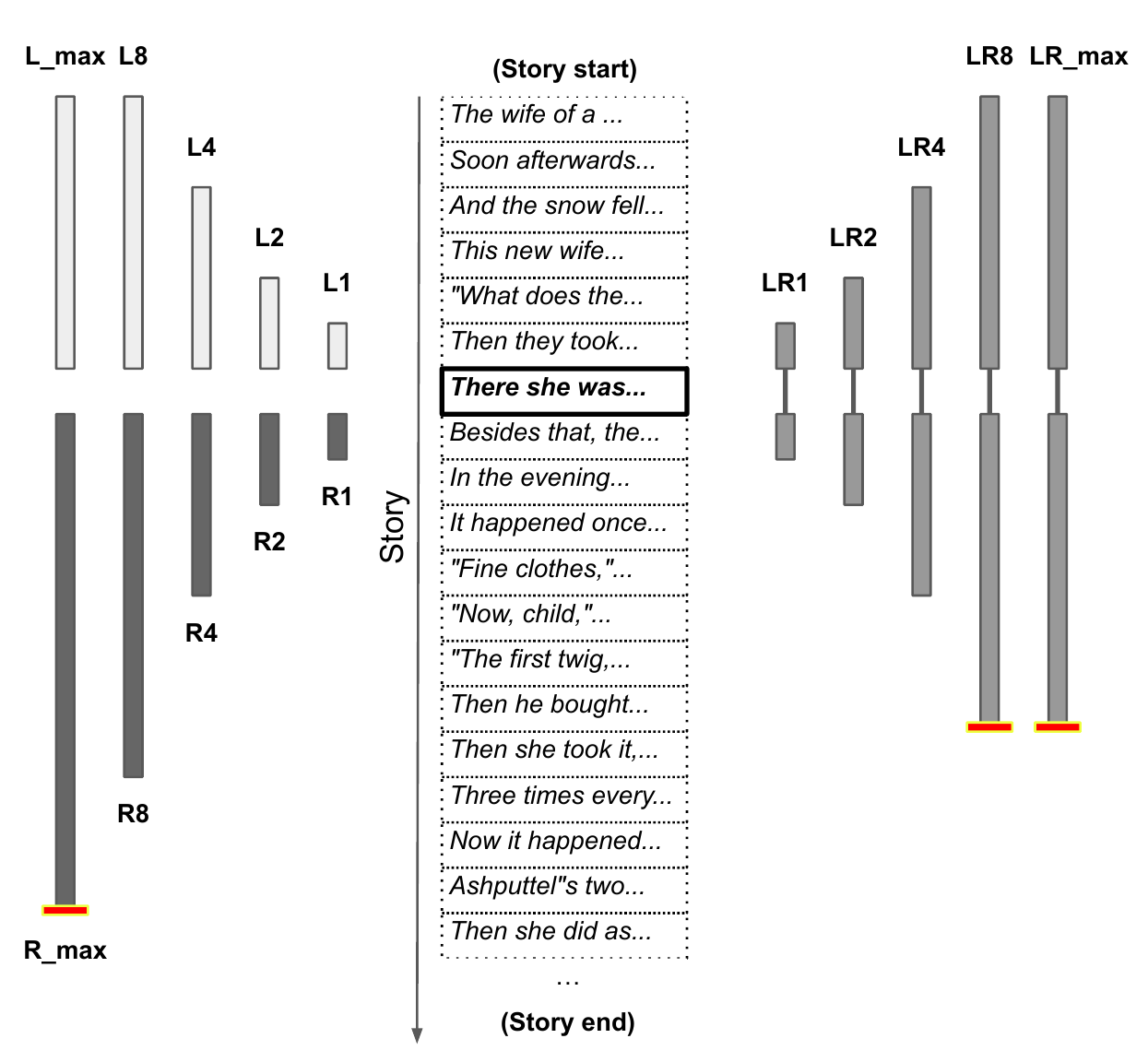}
    \caption{Different strategies for context selection exemplified with the seventh sentence (in bold) of \textit{Grimm's} story \textit{Ashputtel}. The bars span the sentences making up the context when applying the respective context selection approach. Red lines at the end of bars indicate that the context was cut at this sentence in order not to exceed the maximum number of tokens ($512$) in the input. }
    \label{fig:sampling}
\end{figure}

\begin{figure}[h!]
    \centering
    \includegraphics[width=.95\columnwidth]{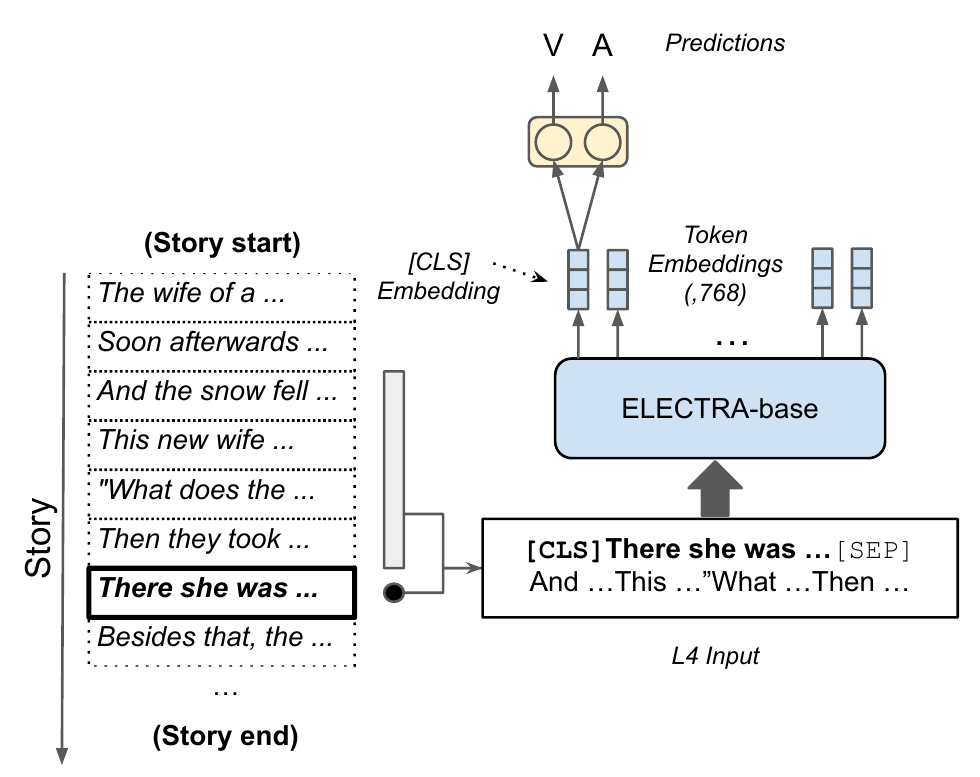}
    \caption{Example for the ELECTRA finetuning approach. Valence (\textit{V}) and arousal (\textit{A}) predictions are obtained for the boldfaced sentence. Here, the \textit{L4} strategy is applied.}
    \label{fig:finetuning}
\end{figure}

The \verb|context| part of the model input is selected via the following policies. We experiment with the left context (\textit{L}) of a sentence, \ie, only those sentences preceding it, its right context (\textit{R}), \ie, only succeeding sentences, and both the left and right context (\textit{LR}) at the same time. In the input for \textit{LR}, we separate the left and right context from each other via the token \verb|#| which does not occur in the data. For all three methods, we vary the number of context sentences. We utilise either $1$, $2$, $4$, $8$, or as many sentences possible ($\_max$) without exceeding the maximum number of input tokens, which is $512$ in the \textsc{ELECTRA} model we employ. Accordingly, every training input is cut if its length would be more than $512$ tokens. When cutting an input, only full sentences are retained. 
For the baseline, denoted $0$, we omit the context. Thus, there are 16 different fine-tuning strategies in total. \Cref{fig:sampling} illustrates the construction of the training examples.  
In~\Cref{fig:finetuning}, a concrete example for an input obtained with the \textit{L4} strategy is given.

For each of the 16 strategies, we fine-tune the model for at most $5$ epochs but abort the training process early if no improvement on the development set is achieved for $2$ consecutive epochs. Adam~\cite{kingma2014adam} is chosen as the optimisation method. The learning rate is set to $5\times10^{-6}$. Dropout with a rate of $0.5$ is applied on the \verb|[CLS]| embedding. Every experiment is repeated with five fixed seeds. Given that we train 16 different configurations with 5 different seeds each, we limit our experiments to the \textit{base} variant of \textsc{ELECTRA}, consisting of 12 Transformer encoder layers that account for about $350$ million parameters overall. In every experiment, we initialise the model with the checkpoint provided by the \textsc{ELECTRA} authors\footnote{\href{https://huggingface.co/google/electra-base-discriminator}{https://huggingface.co/google/electra-base-discriminator}}. The implementation utilises PyTorch and the Hugging Face \verb|transformers| library\footnote{\href{https://huggingface.co/docs/transformers/model_doc/electra}{https://huggingface.co/docs/transformers/model\_doc/electra}}.

All experiments introduced so far are conducted using the data split described in~\Cref{tab:partition}. Afterwards, we take the best-performing fine-tuning strategy for experiments on all author-based splits where each of the three partitions corresponds to one of the three authors. 

\subsection{Context Modelling}\label{ssec:context_modelling}
In the second set of experiments, we freeze the previously fine-tuned \textsc{ELECTRA} models and apply \acp{LSTM} (\Cref{ssec:lstms}) as well as a combination of \acp{LSTM} and Transformer layers (\Cref{ssec:lstm_transformers}) to the sentence embeddings provided by the \textsc{ELECTRA} models. We only make use of the author-independent main data split for these experiments.

\subsubsection{LSTMs}\label{ssec:lstms}
\acp{LSTM} model sequential data in a recurrent manner, \ie, storing latent information from previously read data. %
Therefore, we hope for a better generalisation for modelling the emotions across a whole story. Given one of the 16 fine-tuned models introduced in~\Cref{ssec:finetuning}, we extract the \verb|[CLS]| embeddings per sentence, utilising a sentence's context in the same way the respective \textsc{ELECTRA} model was trained. We then feed these embeddings of consecutive sentences into a bidirectional \ac{LSTM}. The \ac{LSTM} representations for each of the sentences are subsequently passed through a feed-forward layer that projects them into $2$ dimensions. Finally, a Sigmoid activation yields valence and arousal predictions for each of the sentences. The same loss function as in~\Cref{ssec:finetuning} is utilised.

Since there are only $118$ stories in the training partition (cf.~\Cref{tab:partition}) and deep neural networks typically benefit from a large number of training examples, we do not use complete stories as training examples.
Instead, we construct the training data set by employing a sliding window approach to sample passages from stories. The frame size, \ie, the maximum number of sentences per sample, and the step size, \ie, the number of sentences to skip in each sampling step, are hyperparameters which we tune.

We conduct a grid search over the hyperparameters listed in \Cref{tab:hyperparams} and select the model with the best development performance for the evaluation on the test partition. Even though a performance improvement on the development partition is promoted by the sheer number of hyperparameter experiments, we observe a consistently good generalisation to the test data in~\Cref{sec:results}.

\begin{table}[h!]
    \centering\resizebox{1\columnwidth}{!}{
    
    \begin{tabular}{lrr}
          \toprule
         \multicolumn{1}{l}{\textbf{Parameter}} &          
         \multicolumn{1}{r}{\textbf{LSTM}} & 
         \multicolumn{1}{r}{\textbf{LSTM + Tr.}} 
         \\
         \midrule

         \#LSTM layers & \{$1$,$\mathbf{2}$,$4$\} & \{$1$, $\mathbf{2}$\} \\
         
         LSTM hidden size & \{$64$, $128$, $\mathbf{256}$, $512$\} & \{$64$, $128$, $\mathbf{256}$, $512$\} \\

         \#$T_{L}$ & \multicolumn{1}{r}{-} & \{$1$, $\mathbf{2}$, $4$\} \\

         $T_{AW}$ & \multicolumn{1}{r}{-} & \{$\mathbf{1}$, $2$, $4$\} \\ 

         \#$T_{H}$ & \multicolumn{1}{r}{-} & \{$1$, $\mathbf{2}$, $4$\} \\

         Window size & \{$3$, $5$, $\mathbf{10}$, $20$\} & \{$3$, $5$, $\mathbf{10}$, $20$\} \\ 

         Step size & \{$1$, $2$, $\mathbf{4}$\} & \{$1$, $\mathbf{2}$, $4$\} \\ 

         $\eta$ & \{$10^{-5}$, $5\times 10^{-5}$, $\mathbf{10^{-4}}$\} & \{$10^{-5}$, $5\times10^{-5}$, $\mathbf{10^{-4}}$\} \\ 

         \#Epochs (max.) & $\mathbf{20}$ & $\mathbf{20}$ \\ 

         Early Stopping & $\mathbf{5}$ & $\mathbf{5}$ \\

         \bottomrule
         
    \end{tabular}
    }
    \caption{Hyperparameters and their range considered in the hyperparameter search for both the \ac{LSTM} and the \ac{LSTM} + Transformers (\textit{Tr.}) approach. \#$T_L$ denotes the number of Transformer encoder layers, \#$T_{AW}$ is the size of the Transformer self-attention window, and \#$T_H$ the number of self-attention heads. The best configurations are boldfaced.} 
    \label{tab:hyperparams}
\end{table}

\subsubsection{LSTMs + Transformers}\label{ssec:lstm_transformers}
\begin{figure}[h!]
    \centering
    \includegraphics[width=.85\columnwidth]{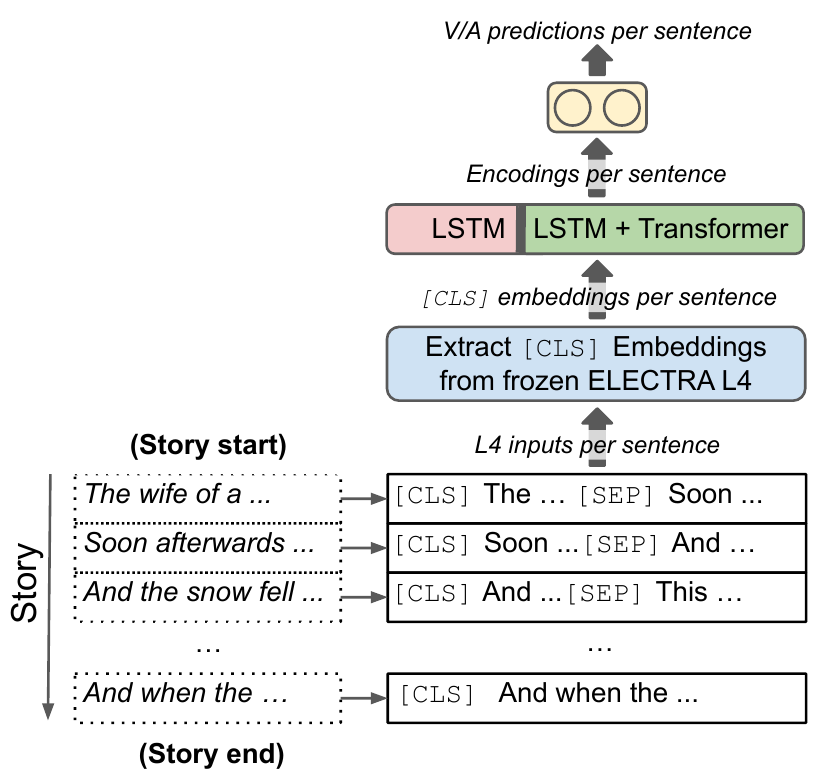}
    \caption{Test time prediction for the story \textit{Ashputtel} with the context modelling approach, exemplified with the \textit{L4} backbone.}
    \label{fig:context_modelling}
\end{figure}

Other than \acp{LSTM}, Transformers~\cite{vaswani2017attention} process sequential data in a non-recurrent way. The self-attention mechanism as a central element of the Transformers architecture allows each element in the sequence to attend to every other element. As most contemporary pretrained language models are Transformer-based (\eg,~\cite{devlin2018bert, lan2019albert, clark2020electra}), we also experiment with Transformers for modelling emotionality.
Identical to the experiments described in~\Cref{ssec:lstms}, we take sequences of consecutive sentences' \verb|[CLS]| token embeddings as our input data. These are first passed through at least one bidirectional \ac{LSTM} layer before the resulting sentence representations are processed by at least one Transformer encoder layer. This, in turn, results in another sequence of sentence embeddings. Analogously to the \ac{LSTM} experiments, the sentence representations are reduced to $2$ dimensions via a feed-forward layer before the Sigmoid function is applied, yielding a valence and arousal prediction for each sentence. At test time, we feed complete stories into the model. Our approach is exemplified in~\Cref{fig:context_modelling}.

We also conducted experiments without the intermediate \ac{LSTM} layer(s) but found that this approach yields results considerably below baseline, which are thus not further discussed here. Moreover, we modify the self-attention masks utilised in the Transformer encoder layers. We found it beneficial for a sentence embedding to only attend to its immediate context. Hence, we experiment with attention masks that constrain an element's attention to the $1$, $2$, and $4$ elements surrounding it to its left and right. Analogously to the \ac{LSTM} experiments, we create training examples with a sliding window approach, for which we also treat the frame and step size as hyperparameters to be tuned. A complete overview of all hyperparameters is provided in~\Cref{tab:hyperparams}.

\section{Results}\label{sec:results}

We report the results on both the main data and author-independent splits in~\Cref{ssec:main_results} and~\Cref{ssec:author_results}, respectively. 
Furthermore, we provide a more fine-grained qualitative analysis of the best model's predictions in~\Cref{ssec:qualitative}. The \ac{CCC} is utilised for evaluating all results. It is calculated over the whole development and test data partition, respectively.

\subsection{Finetuning ELECTRA and Context Modelling}\label{ssec:main_results}

\begin{table*}[h!]
\centering
\begin{tabular}{l|rrrr|rrrr|rrrr}
\toprule
 \multirow{3}{*}{FT strategy}& \multicolumn{4}{c|}{\textbf{FT only [CCC$\uparrow$]}} & \multicolumn{4}{c|}{\textbf{FT+LSTM [CCC$\uparrow$]}} & \multicolumn{4}{c}{\textbf{FT+LSTM+Transformer [CCC$\uparrow$]}}\\
 
 & \multicolumn{2}{c}{Valence} & \multicolumn{2}{c|}{Arousal} & \multicolumn{2}{c}{Valence} & \multicolumn{2}{c|}{Arousal} & \multicolumn{2}{c}{Valence} & \multicolumn{2}{c}{Arousal}  \\

 & \multicolumn{1}{c}{dev} & \multicolumn{1}{c}{test} & \multicolumn{1}{c}{dev} & \multicolumn{1}{c|}{test} & \multicolumn{1}{c}{dev} & \multicolumn{1}{c}{test} & \multicolumn{1}{c}{dev} & \multicolumn{1}{c|}{test} & \multicolumn{1}{c}{dev} & \multicolumn{1}{c}{test} & \multicolumn{1}{c}{dev} & \multicolumn{1}{c}{test} \\ \midrule

0 & .6652 & .6585 & .5283 & .5518 & 
.7143 & .7143 & .5682 & .5848 & 
.7218 & .7163 & .5753 & .5974\\ 

 \midrule

 L1 & .7012 & .6728  & .5633 & .5677 &
 .7248 & .6927 & .5705 & .5830 & 
 .7330 & .7020 & .5839 & .5997 \\ 

 L2 & .7177 & .7017 & .5875 & .6023 & 
 .7215 & .6956 & .5946 & .5839 & 
 .7323 & .7048 & .6096 & .6023 \\ 

 L4 & .7316 & .7119 & .5954 & .6244 & 
 .7292 & .6996 & .5968 & .6096 & 
 .7419 & .7101 & .6074 & .6215 \\ 

 L8 & .7291 & .7148 & \underline{.6053} & .6101 & 
 .7307 & .6962 & \underline{.6061} & .5985 & 
 .7431 & .7106 & \underline{\textbf{.6189}} & .6105
 \\ 

 L\_max & .6966 & .6792 & .5565 & .5699 & 
 .7074 & .6790 & .5556 & .5699 & 
 .7244 & .6954 & .5753 & .5977
 \\ 

 \midrule

 R1 & .6776 & .6797 & .5319 & .5703 & 
 .6922 & .6813 & .5476 & .5635 & 
 .7051 & .6959 & .5590 & .5789
 \\ 

 R2 & .6816 & .6756 & .5314 & .5713 & 
 .7036 & .6893 & .5464 & .5874 & 
 .7174 & .6980 & .5558 & .6005
 \\ 

 R4 & .6916 & .6664 & .5359 & .5658 & 
 .7114 & .6757 & .5496 & .5620 & 
 .7190 & .6845 & .5620 & .5827
 \\ 

 R8 & .6921 & .6698 & .5314 & .5595 & 
 .7086 & .6670 & .5363 & .5429 & 
 .7266 & .6855 & .5486 & .5581 \\ 

 R\_max & .6776 & .6705 & .5391 & .5665 & 
 .6958 & .6801 & .5482 & .5795 & 
 .7095 & .6941 & .5647 & .6002 \\ 

 \midrule

 LR1 & .7084 & .6825 & .5728 & .5890 & 
 .7240 & .6995 & .5736 & .5894 & 
 .7348 & .7059 & .5880 & .6092
 \\ 

 LR2 & .7315 & .7180 & .5895 & .6126 & 
 .7368 & .7080 & .5899 & .5996 & 
 .7459 & .7169 & .5993 & .6141 \\ 

 LR4 & \underline{.7397} & .7327 & .5914 & .6246 & 
 \underline{.7444} & .7202 & .5973 & .6152 & 
 \textbf{\underline{.7500}} & .7338 & .6077 & .6302
 \\ 

 LR8 & .7338 & .7287 & .5931 & .6161 & 
 .7297 & .7237 & .6047 & .6173 & 
 .7392 & .7330 & .6101 & .6287
 \\ 

 LR\_max & .7316 & .7257 & .5946 & .6194 & 
 .7328 & .7149 & .5914 & .6067 & 
 .7408 & .7296 & .5999 & .6173
 \\

\bottomrule

 \end{tabular}\caption{Results for different finetuning (FT) strategies in combination with \acp{LSTM} as well as \ac{LSTM} with Transformers. The results are averaged over 5 fixed seeds. Standard deviations are negligible and thus not given. Overall best results on the development set per prediction target and partition are boldfaced, and the best results for each method are underlined.} \label{tab:main_results}\end{table*}

We fine-tune \textsc{ELECTRA} as described in~\Cref{ssec:finetuning} and, subsequently, freeze the thus obtained models as backbones for the approaches introduced in~\Cref{ssec:lstm_transformers}.
The results for all experiments on the main data split described in~\Cref{tab:partition} are presented in~\Cref{tab:main_results}. 

For the finetuning-only experiments (\textit{FT only}), it can be concluded that considering the context of a sentence always improves the model's performance. All experiments in which context sentences are included in the input lead to better mean \ac{CCC} values than the context-less baselines ($0$) for both valence and arousal, which account for $.6652$ and $.5283$ \ac{CCC} on the development set, respectively. The best configuration for predicting valence on the development set is \textit{LR4} with a \ac{CCC} of $.7397$. Regarding arousal, the best performing strategy is \textit{L8}, achieving $.6053$ on the development set. However, several models,  generalise better to the test set, \eg, \textit{LR4} that yields a \ac{CCC} of $.6246$ for the test set. In general, the fine-tuned models generalise well from the development to the test set. The largest discrepancies observed between development and test \ac{CCC} for one configuration are a drop of about $.0280$ points for valence (\textit{L1}) and an increase of $.0400$ points for arousal (\textit{R2}). 

Strategies only considering the left context consistently outperform those limited to the right context. The best mean result for any right context strategy is $.6921$ \ac{CCC} for valence and $.5391$ for arousal on the development set -- both of which are outperformed by every left context strategy.

The \textit{LR} strategies also prove to be superior over the right context-only strategies in all cases. Comparing the \textit{LR} strategies' results to the left-only results is more difficult. For valence, the best result on the development set is obtained with the \textit{LR4} strategy, while the left-only strategy \textit{L8} proves to perform best for arousal. It can, however, be concluded that the inclusion of both left and right context does not harm the performance.

\begin{figure}
    \centering
    \includegraphics[width=\columnwidth]{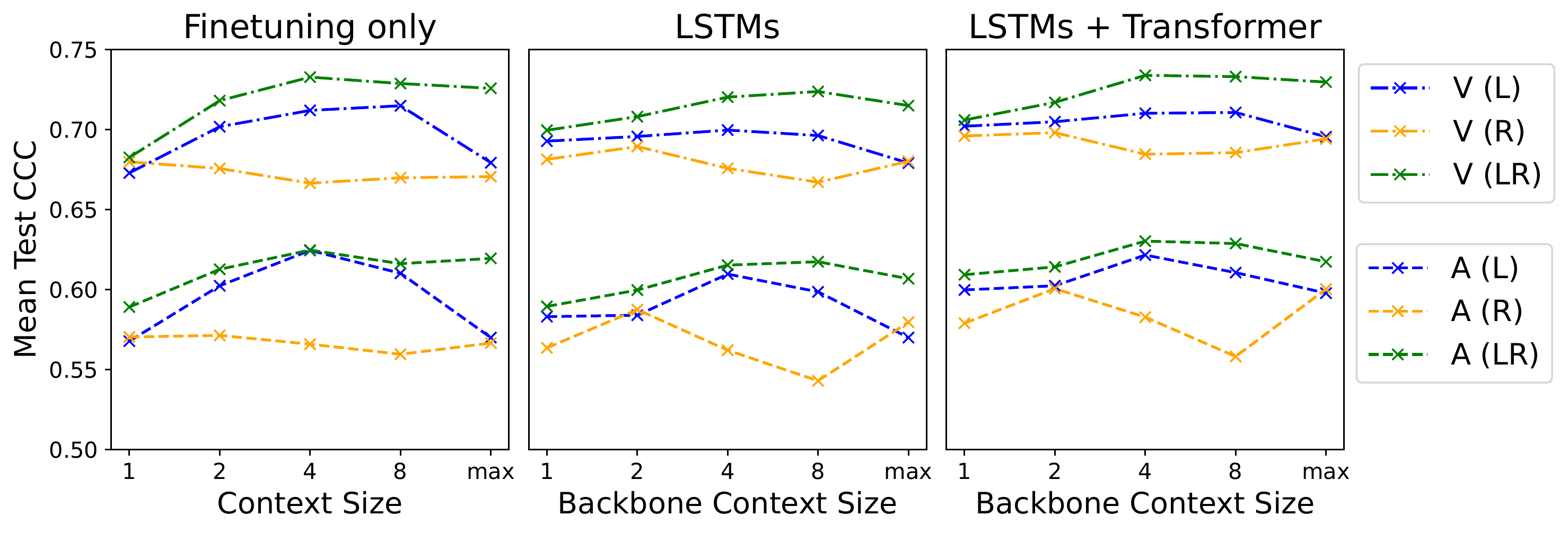}
    \caption{Mean \ac{CCC} values for valence (\textit{V}) and arousal (\textit{A}) on the main split test set obtained with different approaches.}
    \label{fig:contexts}
\end{figure}

Regarding the effect of context window sizes, a few general patterns can be observed. \Cref{fig:contexts} plots the context window sizes against the test set \ac{CCC} results for all three experimental setups. 
In the left context-only configurations, \ac{CCC} values for both labels tend to increase with the context window size until a window size of $4$ or $8$, but always drop for the \textit{\_max} configurations. When utilising only the right context, increasing the context window size rarely leads to better results. Instead, the best results of the right context-only results are usually obtained with a context window size of $1$ or $2$. Similar to the left-context only configurations the results for \textit{LR} configurations typically increase until a context size of $4$ or $8$ sentences to the left and right is reached.

Both additional context modelling approaches (cf.~\Cref{ssec:context_modelling}) mirror the results obtained with the different fine-tuning strategies. There are no notable generalisation problems here, either, with discrepancies between development and test \ac{CCC} results never exceeding $.05$ points. The models built on top of \textsc{ELECTRA} models fine-tuned utilising the right context only can typically not compete with those based on left-context only models. For instance, the best valence \ac{CCC} obtained with \textit{R} backbones on the development set is $.7114$, while with a \textit{L8} backbone \ac{LSTM}, a \ac{CCC} of $.7307$ is achieved. The discrepancy between right context-based \acp{LSTM} and left context-based ones is even more notable for the arousal results, where the best arousal \ac{CCC} on the development set for any \textit{R} strategy is $.5496$, compared to $.6061$ obtained with the \textit{L8} backbone  An analogous argument can be made with respect to the \ac{LSTM} + Transformers experiments. Reflecting the \textsc{ELECTRA} finetuning results, the \ac{LSTM} and \ac{LSTM} + Transformers based on \textit{LR} backbones prove to be superior over those based on models trained with \textit{R} strategies.

Overall, the \ac{LSTM} approach outperforms the fine-tuning method for predicting valence. The best \ac{CCC} value obtained with \acp{LSTM} on the development set is $.7444$, while the fine-tuning approach achieves a \ac{CCC} of $.7397$. For arousal, \acp{LSTM} only slightly surpass the fine-tuning approach with a development arousal \ac{CCC} of $.6061$.
The combination of \acp{LSTM} and Transformers account for the best results overall, with \acp{CCC} of $.7500$ and $.6189$ for valence and arousal on the development set, respectively. For all $16$ backbones, the \ac{LSTM} + Transformer models outperform their \ac{LSTM}-only counterpart for both prediction targets and on both the development and test partitions.

\subsection{Author-independent Results}\label{ssec:author_results}
We opt for the \textit{LR4} strategy to fine-tune \textsc{ELECTRA} on the $9$ author-independent data splits defined by the three authors in the dataset. Yielding the best development valence \ac{CCC} and the best test results for both valence and arousal, \textit{LR4} is arguably the best performing strategy. The results of the author-independent split experiments are provided in~\Cref{tab:author-results}.

\begin{table}[h!]
\centering
\begin{tabular}{lllrrrr}
\toprule 
     \multicolumn{3}{c}{Partition} & \multicolumn{2}{c}{Valence [CCC$\uparrow$]} & \multicolumn{2}{c}{Arousal [CCC$\uparrow$]}  \\ 
     \cmidrule(lr){0-2} \cmidrule(lr){4-5} \cmidrule(lr){6-7}
     \multicolumn{1}{l}{train} & \multicolumn{1}{l}{dev} & \multicolumn{1}{l}{test} & \multicolumn{1}{c}{dev} & \multicolumn{1}{c}{test} & \multicolumn{1}{c}{dev} & \multicolumn{1}{c}{test} \\ \midrule

    Gri &  HCA & Pot & .6899 & .6003 &  .5728 & .5531   \\
    Gri &  Pot & HCA & .6120 & .6856 &  .5548 & .5633  \\
    \midrule
    HCA &  Gri & Pot & .6910 & .6393 &  .5769 & .5903  \\
    HCA &  Pot & Gri & .6343 & .6870 &  .5985 & .5745  \\
    \midrule
    Pot &  Gri & HCA & .5554 & .5840 &  .4387 & .4854  \\
    Pot &  HCA & Gri & .5840 & .5554 &  .4854 & .4387  \\
     
\bottomrule
     
\end{tabular}
\caption{Results for all possible author-independent splits, obtained with the \textit{LR4} finetuning only approach. We report the average \acp{CCC} over $3$ fixed seeds. Authors are abbreviated with \emph{Gri} (\textit{Brothers Grimm}), \emph{HCA} (\textit{Andersen}), and \emph{Pot} (\textit{Potter}). Standard deviations are low and thus omitted.}\label{tab:author-results}
\end{table}

For both valence and arousal, the mean \ac{CCC} values on the development and test data are typically lower than the corresponding ones reported for the main split. No valence result in the author-independent experiments reaches the \ac{CCC} values observed with the main split, namely $.7397$ and $.7327$ for the development and test partition, respectively (cf.~\Cref{tab:main_results}). The highest \ac{CCC} for valence are achieved when generalising from \textit{Andersen} to the \textit{Grimm} and vice versa, with \ac{CCC} values between $.6856$ (training on \textit{Grimm}, testing with \textit{Andersen}) and $.6910$ (training on \textit{Andersen}, \textit{Grimm} as development set). 
 Similarly, the author-independent arousal \ac{CCC} results are lower than those obtained with the main split. Only the arousal \ac{CCC} results for \textit{Potter} yielded by models trained on \textit{Andersen's} stories are comparable to the arousal \ac{CCC} values reported for the main split. For example, a \ac{CCC} of $.5985$ is achieved when \textit{Potter's} stories serve as the development set for the \textit{Andersen}-trained model, while the mean development set \ac{CCC} for arousal in the main split experiments with the \textit{LR4} strategy (cf.~\Cref{tab:main_results}) is $.5914$. 
It is clear from these results that models trained on one particular author cannot always be expected to generalise well to the stories of other authors, indicating that our method tends to adapt to author-specific characteristics. For instance, when training on \textit{Potter's} stories only while utilising \textit{Andersen's} stories as the development set, a valence \ac{CCC} of $.5554$ is obtained for the test set consisting of the \textit{Grimm} brothers' stories. In contrast, when the model is trained on \textit{Andersen's} stories, a valence \ac{CCC} for the \textit{Grimms} test set of $.6870$ is obtained. Similarly, the arousal \acp{CCC} for the \textit{Grimms} test set differ depending on the training data. The corresponding \ac{CCC} values are $.4387$ when training on the \textit{Potter} partition but $.5745$ when employing \textit{Andersen's} stories as training data. Partly, the comparably poor generalisation from \textit{Potter} to the other two authors might be due to the comparably small size of the \textit{Potter} partition, providing the model with less information to learn from in comparison to the other two authors' data. Another aspect is that the topics of the stories by \textit{Andersen} and the \textit{Grimm} brothers differ from those of \textit{Potter's} stories. While the tales of \textit{Andersen} and the \textit{Grimm} brothers typically feature human protagonists, animals -- in particular, rabbits -- are the main characters in \textit{Potter's} works. We empirically underpin this reasoning by the means of topic modelling. First, every story is reduced to its lemmatised, non-stop word nouns via spaCy~\cite{honnibal2020spacy}. Then, the set of each author's stories is fed into a \ac{LDA}~\cite{hoffman2010online, blei2003latent} model, with the number of topics set to $5$. \Cref{tab:lda} displays the resulting top three words per topic and author.

\begin{table}[h!]
    \centering\resizebox{0.85\columnwidth}{!}{
    \begin{tabular}{ll}
    \toprule
         \multicolumn{1}{c}{Author} & \multicolumn{1}{c}{Topics}  \\ \hline 
         \multirow{3}{*}{Grimm} & \{day, water, child\}, \{man, time, day\}, 
         \\ & \{king, father, man\},
          \{fox, princess, soldier\}, \\ & \{child, woman, peasant\} \\ 
          \midrule

          \multirow{3}{*}{HCA} & \{bottle, bird, people\}, \{tree, bird, boy\}, 
         \\ & \{soldier, flower, princess\},
          \{time, day, thing\}, \\ & \{boy, man, child\} \\ 
          \midrule

          \multirow{3}{*}{Potter} & \{shop, water, mouse\}, \{pocket, clothe, coat\}, 
         \\ & \{rabbit, sack, head\},
          \{duchess, house, pie\}, \\ & \{rabbit, pig, house\} \\ 

          \bottomrule
    \end{tabular}}
    \caption{\ac{LDA} results per author. The top three words in each of the five topics per author are listed.}
    \label{tab:lda}
\end{table}

Clearly, several words in \textit{Potter's} topics refer to animals, \eg, \textit{rabbit} and \textit{mouse}. The topics of \textit{Andersen's} and the \textit{Grimm} brothers' stories are, in general, more related to humans, \eg, \textit{child}, \textit{man} and \textit{soldier}.

\subsection{Qualitative Analysis}\label{ssec:qualitative}
For a more detailed analysis, we again focus on the \textsc{ELECTRA} model trained with the \textit{LR4} strategy on the main data split. 

\begin{table}[h!]
    \centering
    \begin{tabular}{lrr}
    \toprule
         CCC$\uparrow$ per story &  \multicolumn{1}{c}{Valence} & \multicolumn{1}{c}{Arousal}\\ \midrule
         Overall & .6636 ($\pm$.1293) & .5563 ($\pm$.1592) \\ 
         Grimm & .6508 ($\pm$.1131) & .5102 ($\pm$.1733) \\ 
         HCA & \textbf{.7218} ($\pm$.1253) & \textbf{.6585} ($\pm$.0789) \\ 
         Potter & .5484 ($\pm$.1179) & .4643 ($\pm$.1063) \\ 
         \bottomrule
    \end{tabular}
    \caption{Statistics on story-wise \ac{CCC} values for predictions on the test data. The predictions were obtained with the best out of $5$ \textit{LR4} models.}
    \label{tab:ccc_per_story}
\end{table}
\Cref{tab:ccc_per_story} provides story-level statistics on the test set predictions of the best \textit{LR4} seed. The author-wise means demonstrate again that our method's generalisation capabilities vary for different authors. For instance, in \textit{Potter's} test stories, a mean valence \ac{CCC} of $.5484$ is obtained, while for \textit{Andersen}, the mean story-wise valence \ac{CCC} is $.7218$. Similar to our reasoning in~\Cref{ssec:author_results}, the comparably low performance for \textit{Potter} can be explained with her stories being underrepresented in the training data (cf.~\Cref{tab:partition}) and featuring topics not to be found in the stories of the other two authors. Furthermore, the model's performance differs considerably with respect to different stories from the same author. This is demonstrated by the high standard deviations, \eg, $.1293$ for valence and $.1592$ for arousal when calculated over all test stories.

\begin{figure}[h!]
    \centering
    \includegraphics[width=\columnwidth]{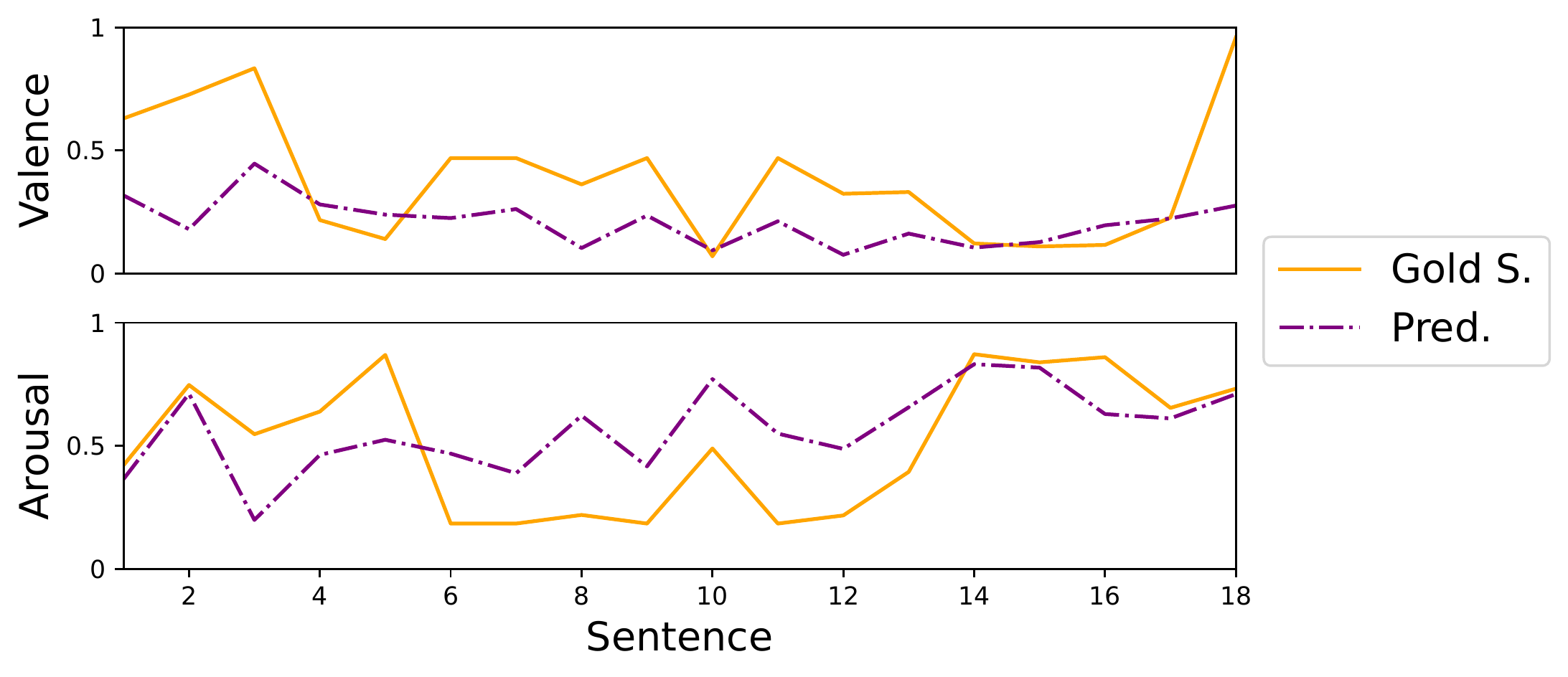}
    \caption{Gold Standards and Predictions of the best \textit{LR4} model for \textit{The Story of Miss Moppet}.}
    \label{fig:pred_mm}
\end{figure}

In~\Cref{fig:pred_mm}, the gold standard and the \textit{L8} model's predictions for \textit{Potter's} test story \textit{The Story of Miss Moppet} are plotted. The corresponding valence \ac{CCC} is $.2663$, while for arousal, a \ac{CCC} of $.4295$ is achieved.

The most remarkable differences between the model's valence predictions and the respective gold standard occur at the story's beginning and end. In the story, a cat named \textit{Miss Moppet} is provoked by a mouse and, subsequently fails to catch it.~\Cref{tab:example_story} lists a selection of sentences from the story along with their respective gold standard and predictions for valence.

\begin{table}[h!]\resizebox{1\columnwidth}{!}{
    \centering
    \begin{tabular}{rlrr}
    \toprule
         \# & Text & V (G) & V (P)  \\ \midrule
         \multirow{2}{*}{2} & This is the Mouse peeping out behind the & \multirow{2}{*}{.7270} & \multirow{2}{*}{.1799} \\
         & cupboard and making fun of Miss Moppet. & & \\

         \multirow{1}{*}{10} & Miss Moppet looks worse and worse. & \multirow{1}{*}{.0700} & \multirow{1}{*}{.0939} \\

         \multirow{2}{*}{18} & He has wriggled out and run away; and he & \multirow{2}{*}{.9590} & \multirow{2}{*}{.2760} \\
         & is dancing a jig on top of the cupboard! & & \\
         \bottomrule        
    \end{tabular}}
    \caption{Selection of sentences from \textit{The Story of Miss Moppet} with their valence gold standard (\textit{V (G)}) and predictions (\textit{V (P)}) of the best \textit{LR4} model. \# denotes the position of the sentence in the story. In sentence $18$, \textit{he} is referring to the mouse.}
    \label{tab:example_story}
\end{table}

Apparently, the annotators perceived the story as rather funny considering the mouse is the hero. In sentences $2$ and $18$, however, it seems that the model classifies the provocative mouse as unpleasant. Sentence $10$ describes the increasingly angry cat. In this obvious example, both model and annotators agree upon a negative mood expression.

\begin{figure}[h!]
    \centering
    \includegraphics[width=\columnwidth]{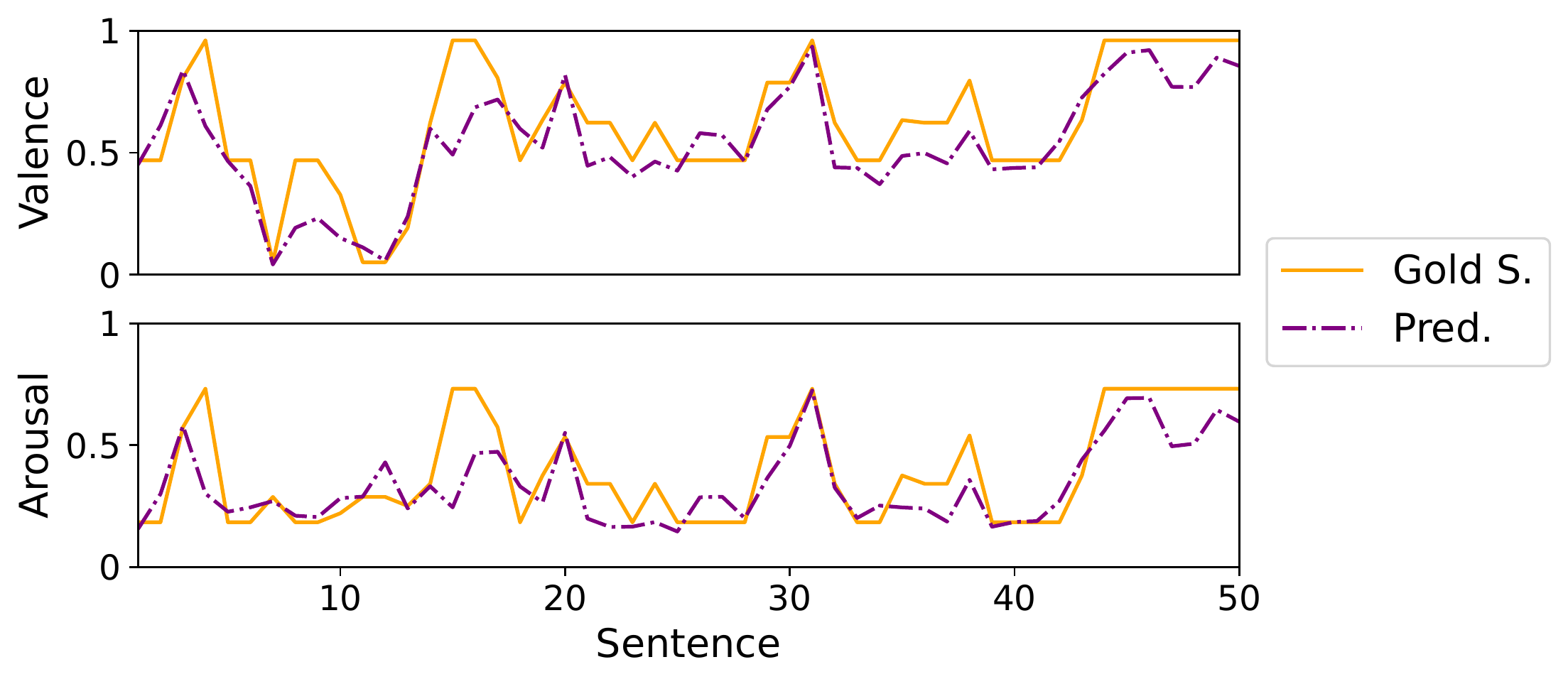}
    \caption{Gold standard and predictions of the best \textit{LR4} model for \textit{popular}.}
    \label{fig:pred_good}
\end{figure}

Other stories in the test set lead to better results than \textit{The Story of Miss Moppet}. To give an example,~\Cref{fig:pred_good} visualises the gold standard and predictions for a story by \textit{Andersen} referred to as \textit{popular} in the dataset.

For \textit{popular}, the model achieves \ac{CCC} values of $.8220$ and $.7376$ for valence and arousal, respectively. These high correlations are clearly shown in the plots, where the gold standard and prediction signals are almost fully aligned.

\section{Conclusions}
In this work, we extended an existing dataset of children's stories~\cite{alm2008affect} with additional annotations and created a continuous valence/arousal gold standard for emotion detection in stories. We employed a variety of Transformer-based methods to model the thus obtained valence/arousal trajectories over the course of stories and provided extensive results. 
We  achieved a \acp{CCC} of up to $.7338$ and $.6302$ for valence and arousal on a held-out test set, respectively, demonstrating the efficacy of our proposed approaches.
Our experiments showed that the integration of the context of a sentence is crucial for this recognition task, in particular, the left context of a sentence, \ie, the sentences preceding it, proved to be more valuable than the right context in general. We observed that adding \ac{LSTM} and Transformer encoder layers following the \textsc{ELECTRA} model could further improve the performance. 
The prediction quality varied considerably for different stories and partitioning strategies (author-dependent or author-independent), demonstrating the complexity of emotion modelling in stories. 
Besides, the proposed Transformer method is resource-intensive when compared to more traditional lexicon-based approaches, limiting its large-scale applicability to very long stories such as novels. 
Another limitation is posed by the static mapping from discrete emotion labels to the valence/arousal space (cf.~\Cref{tab:mapping}). Due to its simplicity, for some instances, this scheme may be too coarse-grained to fully capture the mood transported by a sentence~\cite{barrett2021navigating}.

In future work, the application of a more informed context selection method should be pursued. Such a method could learn to ignore irrelevant neighbouring sentences and dynamically select informative sentences to the left and right. 
Further, personalisation methods (\eg,~\cite{kathan2022personalised}) can help improve the generalisation of author-independent partitions.
Finally, the inclusion of metadata about the authors or even particular stories could increase the emotion recognition performance.

\ifCLASSOPTIONcompsoc
  \section*{Acknowledgments}
\else
  \section*{Acknowledgment}
\fi

This research was partially supported by the Affective Computing \& HCI Innovation Research Lab between Huawei Technologies and the University of Augsburg.

\ifCLASSOPTIONcaptionsoff
  \newpage
\fi

\bibliographystyle{IEEEtranN}
\bibliography{refs}

\begin{thebibliography}{69}
\providecommand{\natexlab}[1]{#1}
\providecommand{\url}[1]{#1}
\csname url@samestyle\endcsname
\providecommand{\newblock}{\relax}
\providecommand{\bibinfo}[2]{#2}
\providecommand{\BIBentrySTDinterwordspacing}{\spaceskip=0pt\relax}
\providecommand{\BIBentryALTinterwordstretchfactor}{4}
\providecommand{\BIBentryALTinterwordspacing}{\spaceskip=\fontdimen2\font plus
\BIBentryALTinterwordstretchfactor\fontdimen3\font minus
  \fontdimen4\font\relax}
\providecommand{\BIBforeignlanguage}[2]{{%
\expandafter\ifx\csname l@#1\endcsname\relax
\typeout{** WARNING: IEEEtranN.bst: No hyphenation pattern has been}%
\typeout{** loaded for the language `#1'. Using the pattern for}%
\typeout{** the default language instead.}%
\else
\language=\csname l@#1\endcsname
\fi
#2}}
\providecommand{\BIBdecl}{\relax}
\BIBdecl

\bibitem[Gottschall(2012)]{gottschall2012storytelling}
J.~Gottschall, \emph{The Storytelling Animal: How Stories Make Us Human}.\hskip
  1em plus 0.5em minus 0.4em\relax Houghton Mifflin Harcourt, 2012.

\bibitem[Maclean et~al.(2015)Maclean, Harvey, Gordon, and
  Shaw]{maclean2015identity}
M.~Maclean, C.~Harvey, J.~Gordon, and E.~Shaw, ``Identity, storytelling and the
  philanthropic journey,'' \emph{human relations}, vol.~68, no.~10, pp.
  1623--1652, 2015.

\bibitem[Humle(2014)]{humle2014remembering}
D.~M. Humle, ``Remembering who we are: Memories of identity through
  storytelling,'' \emph{Tamara Journal of Critical Organisation Inquiry},
  vol.~12, no.~3, p.~11, 2014.

\bibitem[Polletta et~al.(2011)Polletta, Chen, Gardner, and
  Motes]{polletta2011sociology}
F.~Polletta, P.~C.~B. Chen, B.~G. Gardner, and A.~Motes, ``The sociology of
  storytelling,'' \emph{Annual review of sociology}, vol.~37, no.~1, pp.
  109--130, 2011.

\bibitem[Sunderland(2017)]{sunderland2017using}
M.~Sunderland, \emph{Using Story Telling as a Therapeutic Tool with
  Children}.\hskip 1em plus 0.5em minus 0.4em\relax Routledge, 2017.

\bibitem[Boyd(2010)]{boyd2010origin}
B.~Boyd, \emph{On the Origin of Stories: Evolution, Cognition, and
  Fiction}.\hskip 1em plus 0.5em minus 0.4em\relax Harvard University Press,
  2010.

\bibitem[Burke(2015)]{burke2015neuroaesthetics}
M.~Burke, ``The neuroaesthetics of prose fiction: Pitfalls, parameters and
  prospects,'' \emph{Frontiers in Human Neuroscience}, vol.~9, p. 442, 2015.

\bibitem[Palombini(2017)]{palombini2017storytelling}
A.~Palombini, ``Storytelling and telling history. towards a grammar of
  narratives for cultural heritage dissemination in the digital era,''
  \emph{Journal of cultural heritage}, vol.~24, pp. 134--139, 2017.

\bibitem[Anderson(2010)]{anderson2010storytelling}
K.~E. Anderson, ``Storytelling,'' in \emph{21st Century Anthropology: a
  Reference Handbook}.\hskip 1em plus 0.5em minus 0.4em\relax SAGE, 2010.

\bibitem[Hogan(2011)]{hogan2011affective}
P.~C. Hogan, \emph{Affective Narratology: The Emotional Structure of
  Stories}.\hskip 1em plus 0.5em minus 0.4em\relax U of Nebraska Press, 2011.

\bibitem[Reagan et~al.(2016)Reagan, Mitchell, Kiley, Danforth, and
  Dodds]{reagan2016emotional}
A.~J. Reagan, L.~Mitchell, D.~Kiley, C.~M. Danforth, and P.~S. Dodds, ``The
  emotional arcs of stories are dominated by six basic shapes,'' \emph{EPJ Data
  Science}, vol.~5, no.~1, pp. 1--12, 2016.

\bibitem[Somasundaran et~al.(2020)Somasundaran, Chen, and
  Flor]{somasundaran2020emotion}
S.~Somasundaran, X.~Chen, and M.~Flor, ``Emotion arcs of student narratives,''
  in \emph{Proceedings of the First Joint Workshop on Narrative Understanding,
  Storylines, and Events}, 2020, pp. 97--107.

\bibitem[Agrawal and An(2012)]{agrawal2012unsupervised}
A.~Agrawal and A.~An, ``Unsupervised emotion detection from text using semantic
  and syntactic relations,'' in \emph{2012 IEEE/WIC/ACM International
  Conferences on Web Intelligence and Intelligent Agent Technology},
  vol.~1.\hskip 1em plus 0.5em minus 0.4em\relax IEEE, 2012, pp. 346--353.

\bibitem[Batbaatar et~al.(2019)Batbaatar, Li, and Ryu]{batbaatar2019semantic}
E.~Batbaatar, M.~Li, and K.~H. Ryu, ``Semantic-emotion neural network for
  emotion recognition from text,'' \emph{IEEE Access}, vol.~7, pp.
  111\,866--111\,878, 2019.

\bibitem[Vaswani et~al.(2017)Vaswani, Shazeer, Parmar, Uszkoreit, Jones, Gomez,
  Kaiser, and Polosukhin]{vaswani2017attention}
A.~Vaswani, N.~Shazeer, N.~Parmar, J.~Uszkoreit, L.~Jones, A.~N. Gomez,
  {\L}.~Kaiser, and I.~Polosukhin, ``Attention is all you need,''
  \emph{Advances in neural information processing systems}, vol.~30, 2017.

\bibitem[Devlin et~al.(2018)Devlin, Chang, Lee, and Toutanova]{devlin2018bert}
J.~Devlin, M.-W. Chang, K.~Lee, and K.~Toutanova, ``Bert: Pre-training of deep
  bidirectional transformers for language understanding,'' \emph{arXiv preprint
  arXiv:1810.04805}, 2018.

\bibitem[Jiang et~al.(2019)Jiang, He, Chen, Liu, Gao, and Zhao]{jiang2019smart}
H.~Jiang, P.~He, W.~Chen, X.~Liu, J.~Gao, and T.~Zhao, ``Smart: Robust and
  efficient fine-tuning for pre-trained natural language models through
  principled regularized optimization,'' \emph{arXiv preprint
  arXiv:1911.03437}, 2019.

\bibitem[Yang et~al.(2019)Yang, Dai, Yang, Carbonell, Salakhutdinov, and
  Le]{yang2019xlnet}
Z.~Yang, Z.~Dai, Y.~Yang, J.~Carbonell, R.~R. Salakhutdinov, and Q.~V. Le,
  ``Xlnet: Generalized autoregressive pretraining for language understanding,''
  \emph{Advances in neural information processing systems}, vol.~32, 2019.

\bibitem[Weller and Seppi(2019)]{weller2019humor}
O.~Weller and K.~Seppi, ``Humor detection: A transformer gets the last laugh,''
  \emph{arXiv preprint arXiv:1909.00252}, 2019.

\bibitem[Christ et~al.(2022{\natexlab{a}})Christ, Amiriparian, Kathan,
  M{\"u}ller, K{\"o}nig, and Schuller]{christ2022multimodal}
L.~Christ, S.~Amiriparian, A.~Kathan, N.~M{\"u}ller, A.~K{\"o}nig, and B.~W.
  Schuller, ``Multimodal prediction of spontaneous humour: A novel dataset and
  first results,'' \emph{arXiv preprint arXiv:2209.14272}, 2022.

\bibitem[Acheampong et~al.(2021)Acheampong, Nunoo-Mensah, and
  Chen]{acheampong2021transformer}
F.~A. Acheampong, H.~Nunoo-Mensah, and W.~Chen, ``Transformer models for
  text-based emotion detection: a review of bert-based approaches,''
  \emph{Artificial Intelligence Review}, vol.~54, no.~8, pp. 5789--5829, 2021.

\bibitem[Triantafyllopoulos et~al.(2022)Triantafyllopoulos, Schuller,
  {\.I}ymen, Sezgin, He, Yang, Tzirakis, Liu, Mertes, Andr{\'e},
  et~al.]{triantafyllopoulos2022overview}
A.~Triantafyllopoulos, B.~W. Schuller, G.~{\.I}ymen, M.~Sezgin, X.~He, Z.~Yang,
  P.~Tzirakis, S.~Liu, S.~Mertes, E.~Andr{\'e} \emph{et~al.}, ``An overview of
  affective speech synthesis and conversion in the deep learning era,''
  \emph{arXiv preprint arXiv:2210.03538}, 2022.

\bibitem[Lugrin et~al.(2010)Lugrin, Cavazza, Pizzi, Vogt, and
  Andr{\'e}]{lugrin2010exploring}
J.-L. Lugrin, M.~Cavazza, D.~Pizzi, T.~Vogt, and E.~Andr{\'e}, ``Exploring the
  usability of immersive interactive storytelling,'' in \emph{Proceedings of
  the 17th ACM symposium on virtual reality software and technology}, 2010, pp.
  103--110.

\bibitem[Eisenreich et~al.(2014)Eisenreich, Ott, S{\"u}{\ss}dorf, Willms, and
  Declerck]{eisenreich2014tale}
C.~Eisenreich, J.~Ott, T.~S{\"u}{\ss}dorf, C.~Willms, and T.~Declerck, ``From
  tale to speech: Ontology-based emotion and dialogue annotation of fairy tales
  with a tts output.'' in \emph{ISWC-PD'14: Proceedings of the 2014
  International Conference on Posters \& Demonstrations Track - Volume 1272},
  2014.

\bibitem[Alabdulkarim et~al.(2021)Alabdulkarim, Li, and
  Peng]{alabdulkarim2021automatic}
A.~Alabdulkarim, S.~Li, and X.~Peng, ``Automatic story generation: Challenges
  and attempts,'' \emph{arXiv preprint arXiv:2102.12634}, 2021.

\bibitem[Kim and Klinger(2018{\natexlab{a}})]{kim2018survey}
E.~Kim and R.~Klinger, ``A survey on sentiment and emotion analysis for
  computational literary studies,'' \emph{arXiv preprint arXiv:1808.03137},
  2018.

\bibitem[Mani(2014)]{mani2014computational}
I.~Mani, ``Computational narratology,'' \emph{Handbook of narratology}, pp.
  84--92, 2014.

\bibitem[Piper et~al.(2021)Piper, So, and Bamman]{piper2021narrative}
A.~Piper, R.~J. So, and D.~Bamman, ``Narrative theory for computational
  narrative understanding,'' in \emph{Proceedings of the 2021 Conference on
  Empirical Methods in Natural Language Processing}, 2021, pp. 298--311.

\bibitem[Alm(2008)]{alm2008affect}
E.~C.~O. Alm, \emph{Affect in Text and Speech}.\hskip 1em plus 0.5em minus
  0.4em\relax University of Illinois at Urbana-Champaign, 2008.

\bibitem[Russell(1980)]{russell1980circumplex}
J.~A. Russell, ``A circumplex model of affect.'' \emph{Journal of personality
  and social psychology}, vol.~39, no.~6, p. 1161, 1980.

\bibitem[Mohammad and Turney(2013)]{mohammad2013crowdsourcing}
S.~M. Mohammad and P.~D. Turney, ``Crowdsourcing a word--emotion association
  lexicon,'' \emph{Computational intelligence}, vol.~29, no.~3, pp. 436--465,
  2013.

\bibitem[Kim et~al.(2017)Kim, Pad{\'o}, and Klinger]{kim2017prototypical}
E.~Kim, S.~Pad{\'o}, and R.~Klinger, ``Prototypical emotion developments in
  literary genres,'' in \emph{Proceedings of the Joint SIGHUM Workshop on
  Computational Linguistics for Cultural Heritage, Social Sciences, Humanities
  and Literature}, 2017, pp. 17--26.

\bibitem[Mohammad(2012)]{mohammad2012once}
S.~M. Mohammad, ``From once upon a time to happily ever after: Tracking
  emotions in mail and books,'' \emph{Decision Support Systems}, vol.~53,
  no.~4, pp. 730--741, 2012.

\bibitem[Strapparava et~al.(2004)Strapparava, Valitutti,
  et~al.]{strapparava2004wordnet}
C.~Strapparava, A.~Valitutti \emph{et~al.}, ``Wordnet-affect: an affective
  extension of wordnet,'' in \emph{Lrec}, vol.~4, no. 1083-1086.\hskip 1em plus
  0.5em minus 0.4em\relax Lisbon, Portugal, 2004, p.~40.

\bibitem[Mac~Kim et~al.(2010)Mac~Kim, Valitutti, and Calvo]{mac2010evaluation}
S.~Mac~Kim, A.~Valitutti, and R.~A. Calvo, ``Evaluation of unsupervised emotion
  models to textual affect recognition,'' in \emph{Proceedings of the NAACL HLT
  2010 Workshop on Computational Approaches to Analysis and Generation of
  Emotion in Text}, 2010, pp. 62--70.

\bibitem[Zad and Finlayson(2020)]{zad2020systematic}
S.~Zad and M.~Finlayson, ``Systematic evaluation of a framework for
  unsupervised emotion recognition for narrative text,'' in \emph{Proceedings
  of the First Joint Workshop on Narrative Understanding, Storylines, and
  Events}, 2020, pp. 26--37.

\bibitem[Elsner(2012)]{elsner2012character}
M.~Elsner, ``Character-based kernels for novelistic plot structure,'' in
  \emph{Proceedings of the 13th Conference of the European Chapter of the
  Association for Computational Linguistics}, 2012, pp. 634--644.

\bibitem[Wilson et~al.(2005)Wilson, Wiebe, and Hoffmann]{wilson2005recognizing}
T.~Wilson, J.~Wiebe, and P.~Hoffmann, ``Recognizing contextual polarity in
  phrase-level sentiment analysis,'' in \emph{Proceedings of human language
  technology conference and conference on empirical methods in natural language
  processing}, 2005, pp. 347--354.

\bibitem[Yavuz et~al.(2020)Yavuz, Monti, Dell'Orletta, and
  Tamburini]{yavuz2020analyses}
M.~C. Yavuz, J.~Monti, F.~Dell'Orletta, and F.~Tamburini, ``Analyses of
  character emotions in dramatic works by using emolex unigrams.'' in
  \emph{CLiC-it}, 2020.

\bibitem[Mori et~al.(2019)Mori, Yamane, Ushiku, and Harada]{mori2019narratives}
Y.~Mori, H.~Yamane, Y.~Ushiku, and T.~Harada, ``How narratives move your mind:
  A corpus of shared-character stories for connecting emotional flow and
  interestingness,'' \emph{Information Processing \& Management}, vol.~56,
  no.~5, pp. 1865--1879, 2019.

\bibitem[Liu et~al.(2019)Liu, Osama, and De~Andrade]{liu2019dens}
C.~Liu, M.~Osama, and A.~De~Andrade, ``Dens: A dataset for multi-class emotion
  analysis,'' \emph{arXiv preprint arXiv:1910.11769}, 2019.

\bibitem[Kim and Klinger(2018{\natexlab{b}})]{kim2018feels}
E.~Kim and R.~Klinger, ``Who feels what and why? annotation of a literature
  corpus with semantic roles of emotions,'' in \emph{Proceedings of the 27th
  International Conference on Computational Linguistics}, 2018, pp. 1345--1359.

\bibitem[Kim and Klinger(2019{\natexlab{a}})]{kim2019frowning}
------, ``Frowning frodo, wincing leia, and a seriously great friendship:
  Learning to classify emotional relationships of fictional characters,''
  \emph{arXiv preprint arXiv:1903.12453}, 2019.

\bibitem[Kim and Klinger(2019{\natexlab{b}})]{kim2019analysis}
------, ``An analysis of emotion communication channels in fan fiction: Towards
  emotional storytelling,'' \emph{arXiv preprint arXiv:1906.02402}, 2019.

\bibitem[Ong et~al.(2019)Ong, Wu, Tan, Reddan, Kahhale, Mattek, and
  Zaki]{ong2019modeling}
D.~C. Ong, Z.~Wu, Z.-X. Tan, M.~Reddan, I.~Kahhale, A.~Mattek, and J.~Zaki,
  ``Modeling emotion in complex stories: The stanford emotional narratives
  dataset,'' \emph{IEEE Transactions on Affective Computing}, vol.~12, no.~3,
  pp. 579--594, 2019.

\bibitem[Wu et~al.(2019)Wu, Zhang, Zhi-Xuan, Zaki, and Ong]{wu2019attending}
Z.~Wu, X.~Zhang, T.~Zhi-Xuan, J.~Zaki, and D.~C. Ong, ``Attending to emotional
  narratives,'' in \emph{2019 8th International Conference on Affective
  Computing and Intelligent Interaction (ACII)}.\hskip 1em plus 0.5em minus
  0.4em\relax IEEE, 2019, pp. 648--654.

\bibitem[Rathner et~al.(2018)Rathner, Terhorst, Cummins, Schuller, and
  Baumeister]{rathner2018state}
E.-M. Rathner, Y.~Terhorst, N.~Cummins, B.~Schuller, and H.~Baumeister, ``State
  of mind: Classification through self-reported affect and word use in
  speech,'' \emph{19th Annual Conference of the International Speech
  Communication Association (INTERSPEECH 2018)}, 2018.

\bibitem[Schuller et~al.(2018)Schuller, Steidl, Batliner, Marschik, Baumeister,
  Dong, Hantke, Pokorny, Rathner, Bartl-Pokorny,
  et~al.]{schuller2018interspeech}
B.~Schuller, S.~Steidl, A.~Batliner, P.~B. Marschik, H.~Baumeister, F.~Dong,
  S.~Hantke, F.~B. Pokorny, E.-M. Rathner, K.~D. Bartl-Pokorny \emph{et~al.},
  ``The interspeech 2018 computational paralinguistics challenge: Atypical \&
  self-assessed affect, crying \& heart beats,'' \emph{19th Annual Conference
  of the International Speech Communication Association (INTERSPEECH 2018)},
  2018.

\bibitem[Stappen et~al.(2019)Stappen, Cummins, Me{\ss}ner, Baumeister, Dineley,
  and Schuller]{stappen2019context}
L.~Stappen, N.~Cummins, E.-M. Me{\ss}ner, H.~Baumeister, J.~Dineley, and
  B.~Schuller, ``Context modelling using hierarchical attention networks for
  sentiment and self-assessed emotion detection in spoken narratives,'' in
  \emph{ICASSP 2019-2019 IEEE International Conference on Acoustics, Speech and
  Signal Processing (ICASSP)}.\hskip 1em plus 0.5em minus 0.4em\relax IEEE,
  2019, pp. 6680--6684.

\bibitem[Alm and Sproat(2005)]{alm2005emotional}
C.~O. Alm and R.~Sproat, ``Emotional sequencing and development in fairy
  tales,'' in \emph{International Conference on Affective Computing and
  Intelligent Interaction}.\hskip 1em plus 0.5em minus 0.4em\relax Springer,
  2005, pp. 668--674.

\bibitem[Alm et~al.(2005)Alm, Roth, and Sproat]{alm2005emotions}
C.~O. Alm, D.~Roth, and R.~Sproat, ``Emotions from text: machine learning for
  text-based emotion prediction,'' in \emph{Proceedings of human language
  technology conference and conference on empirical methods in natural language
  processing}, 2005, pp. 579--586.

\bibitem[Udochukwu and He(2015)]{udochukwu2015rule}
O.~Udochukwu and Y.~He, ``A rule-based approach to implicit emotion detection
  in text,'' in \emph{International Conference on Applications of Natural
  Language to Information Systems}.\hskip 1em plus 0.5em minus 0.4em\relax
  Springer, 2015, pp. 197--203.

\bibitem[Grimm and Kroschel(2005)]{grimm2005evaluation}
M.~Grimm and K.~Kroschel, ``Evaluation of natural emotions using self
  assessment manikins,'' in \emph{IEEE Workshop on Automatic Speech Recognition
  and Understanding, 2005.}\hskip 1em plus 0.5em minus 0.4em\relax IEEE, 2005,
  pp. 381--385.

\bibitem[Susanto et~al.(2020)Susanto, Livingstone, Ng, and
  Cambria]{susanto2020hourglass}
Y.~Susanto, A.~G. Livingstone, B.~C. Ng, and E.~Cambria, ``The hourglass model
  revisited,'' \emph{IEEE Intelligent Systems}, vol.~35, no.~5, pp. 96--102,
  2020.

\bibitem[Ortony(2022)]{ortony2022all}
A.~Ortony, ``Are all “basic emotions” emotions? a problem for the (basic)
  emotions construct,'' \emph{Perspectives on Psychological Science}, vol.~17,
  no.~1, pp. 41--61, 2022.

\bibitem[Ringeval et~al.(2019)Ringeval, Schuller, Valstar, Cummins, Cowie,
  Tavabi, Schmitt, Alisamir, Amiriparian, Messner, et~al.]{ringeval2019avec}
F.~Ringeval, B.~Schuller, M.~Valstar, N.~Cummins, R.~Cowie, L.~Tavabi,
  M.~Schmitt, S.~Alisamir, S.~Amiriparian, E.-M. Messner \emph{et~al.}, ``Avec
  2019 workshop and challenge: State-of-mind, detecting depression with ai, and
  cross-cultural affect recognition,'' in \emph{Proceedings of the 9th
  International on Audio/visual Emotion Challenge and Workshop}, 2019, pp.
  3--12.

\bibitem[Stappen et~al.(2021)Stappen, Baird, Christ, Schumann, Sertolli,
  Messner, Cambria, Zhao, and Schuller]{stappen2021muse}
L.~Stappen, A.~Baird, L.~Christ, L.~Schumann, B.~Sertolli, E.-M. Messner,
  E.~Cambria, G.~Zhao, and B.~W. Schuller, ``The muse 2021 multimodal sentiment
  analysis challenge: Sentiment, emotion, physiological-emotion, and stress,''
  in \emph{Proceedings of the 2nd on Multimodal Sentiment Analysis Challenge},
  2021, pp. 5--14.

\bibitem[Christ et~al.(2022{\natexlab{b}})Christ, Amiriparian, Baird, Tzirakis,
  Kathan, Müller, Stappen, Meßner, König, Cowen, Cambria, and
  Schuller]{Christ22-TM2}
L.~Christ, S.~Amiriparian, A.~Baird, P.~Tzirakis, A.~Kathan, N.~Müller,
  L.~Stappen, E.-M. Meßner, A.~König, A.~Cowen, E.~Cambria, and B.~W.
  Schuller, ``The muse 2022 multimodal sentiment analysis challenge: Humor,
  emotional reactions, and stress,'' in \emph{MuSe'22: Proceedings of the 3rd
  Multimodal Sentiment Analysis Workshop and Challenge}.\hskip 1em plus 0.5em
  minus 0.4em\relax Lisbon, Portugal: Association for Computing Machinery,
  2022, pp. 5--14, co-located with ACM Multimedia 2022.

\bibitem[Park et~al.(2021)Park, Kim, Ye, Jeon, Park, and
  Oh]{park-etal-2021-dimensional}
\BIBentryALTinterwordspacing
S.~Park, J.~Kim, S.~Ye, J.~Jeon, H.~Y. Park, and A.~Oh, ``Dimensional emotion
  detection from categorical emotion,'' in \emph{Proceedings of the 2021
  Conference on Empirical Methods in Natural Language Processing}.\hskip 1em
  plus 0.5em minus 0.4em\relax Online and Punta Cana, Dominican Republic:
  Association for Computational Linguistics, Nov. 2021, pp. 4367--4380.
  [Online]. Available: \url{https://aclanthology.org/2021.emnlp-main.358}
\BIBentrySTDinterwordspacing

\bibitem[Mohammad(2018)]{mohammad2018obtaining}
S.~Mohammad, ``Obtaining reliable human ratings of valence, arousal, and
  dominance for 20,000 english words,'' in \emph{Proceedings of the 56th annual
  meeting of the association for computational linguistics (volume 1: Long
  papers)}, 2018, pp. 174--184.

\bibitem[Lan et~al.(2019)Lan, Chen, Goodman, Gimpel, Sharma, and
  Soricut]{lan2019albert}
Z.~Lan, M.~Chen, S.~Goodman, K.~Gimpel, P.~Sharma, and R.~Soricut, ``Albert: A
  lite bert for self-supervised learning of language representations,''
  \emph{arXiv preprint arXiv:1909.11942}, 2019.

\bibitem[Clark et~al.(2020)Clark, Luong, Le, and Manning]{clark2020electra}
K.~Clark, M.-T. Luong, Q.~V. Le, and C.~D. Manning, ``Electra: Pre-training
  text encoders as discriminators rather than generators,'' \emph{arXiv
  preprint arXiv:2003.10555}, 2020.

\bibitem[Kim et~al.(2020)Kim, Ko, Song, Jang, and
  Hong]{kim-etal-2020-contextual}
\BIBentryALTinterwordspacing
J.~Kim, H.~Ko, S.~Song, S.~Jang, and J.~Hong, ``Contextual augmentation of
  pretrained language models for emotion recognition in conversations,'' in
  \emph{Proceedings of the Third Workshop on Computational Modeling of People's
  Opinions, Personality, and Emotion's in Social Media}.\hskip 1em plus 0.5em
  minus 0.4em\relax Barcelona, Spain (Online): Association for Computational
  Linguistics, Dec. 2020, pp. 64--73. [Online]. Available:
  \url{https://aclanthology.org/2020.peoples-1.7}
\BIBentrySTDinterwordspacing

\bibitem[Kingma and Ba(2014)]{kingma2014adam}
D.~P. Kingma and J.~Ba, ``Adam: A method for stochastic optimization,''
  \emph{arXiv preprint arXiv:1412.6980}, 2014.

\bibitem[Honnibal et~al.(2020)Honnibal, Montani, Van~Landeghem, and
  Boyd]{honnibal2020spacy}
M.~Honnibal, I.~Montani, S.~Van~Landeghem, and A.~Boyd, ``spacy:
  Industrial-strength natural language processing in python,'' 2020.

\bibitem[Hoffman et~al.(2010)Hoffman, Bach, and Blei]{hoffman2010online}
M.~Hoffman, F.~Bach, and D.~Blei, ``Online learning for latent dirichlet
  allocation,'' \emph{advances in neural information processing systems},
  vol.~23, 2010.

\bibitem[Blei et~al.(2003)Blei, Ng, and Jordan]{blei2003latent}
D.~M. Blei, A.~Y. Ng, and M.~I. Jordan, ``Latent dirichlet allocation,''
  \emph{Journal of machine Learning research}, vol.~3, no. Jan, pp. 993--1022,
  2003.

\bibitem[Barrett and Westlin(2021)]{barrett2021navigating}
L.~F. Barrett and C.~Westlin, ``Navigating the science of emotion,'' in
  \emph{Emotion measurement}.\hskip 1em plus 0.5em minus 0.4em\relax Elsevier,
  2021, pp. 39--84.

\bibitem[Kathan et~al.(2022)Kathan, Amiriparian, Christ, Triantafyllopoulos,
  M{\"u}ller, K{\"o}nig, and Schuller]{kathan2022personalised}
A.~Kathan, S.~Amiriparian, L.~Christ, A.~Triantafyllopoulos, N.~M{\"u}ller,
  A.~K{\"o}nig, and B.~W. Schuller, ``A personalised approach to audiovisual
  humour recognition and its individual-level fairness,'' in \emph{Proceedings
  of the 3rd International on Multimodal Sentiment Analysis Workshop and
  Challenge}, 2022, pp. 29--36.

\end{thebibliography}

\begin{IEEEbiography}[{\includegraphics[width=1in,height=1.25in,clip,keepaspectratio]{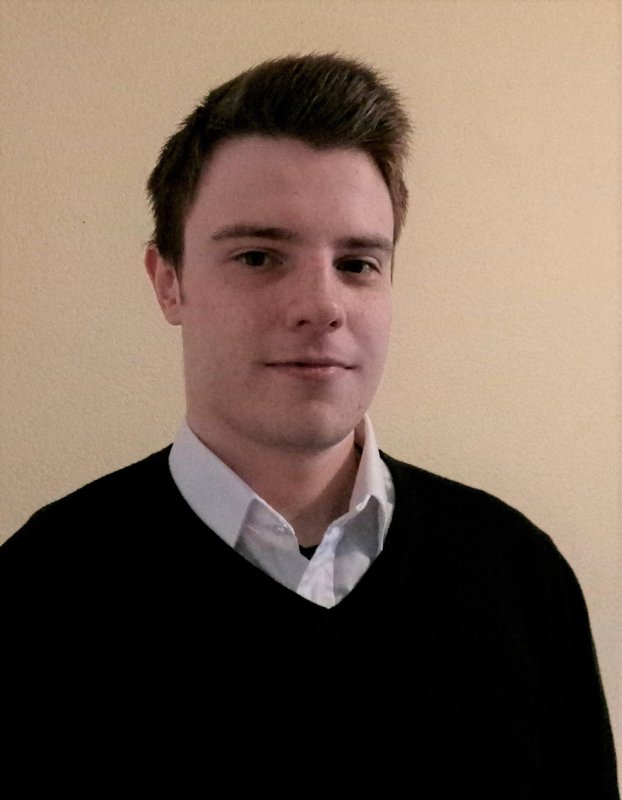}}]{Lukas Christ}
 received his Master's degree in Computer Science at the University of Leipzig in 2020. He is currently a PhD candidate at the Chair of Embedded Intelligence for Health Care and Wellbeing at the University of Augsburg, Germany. His main research interests are natural language processing and multimodal machine learning in the context of affective computing. 
\end{IEEEbiography}	

\begin{IEEEbiography}[{\includegraphics[width=1in,height=1.25in,clip,keepaspectratio]{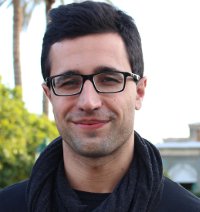}}]{Shahin Amiriparian}
 received his Doctoral degree with the highest honours (summa cum laude) from the Technical University of Munich, Germany in 2019. Currently, he is a postdoctoral researcher at the Chair of Embedded Intelligence for Health Care and Wellbeing, University of Augsburg, Germany. His main research focus is deep learning, unsupervised representation learning, and transfer learning for machine perception, affective computing, and audio understanding.
\end{IEEEbiography}	

\begin{IEEEbiography}[{\includegraphics[width=1in,height=1.25in,clip,keepaspectratio]{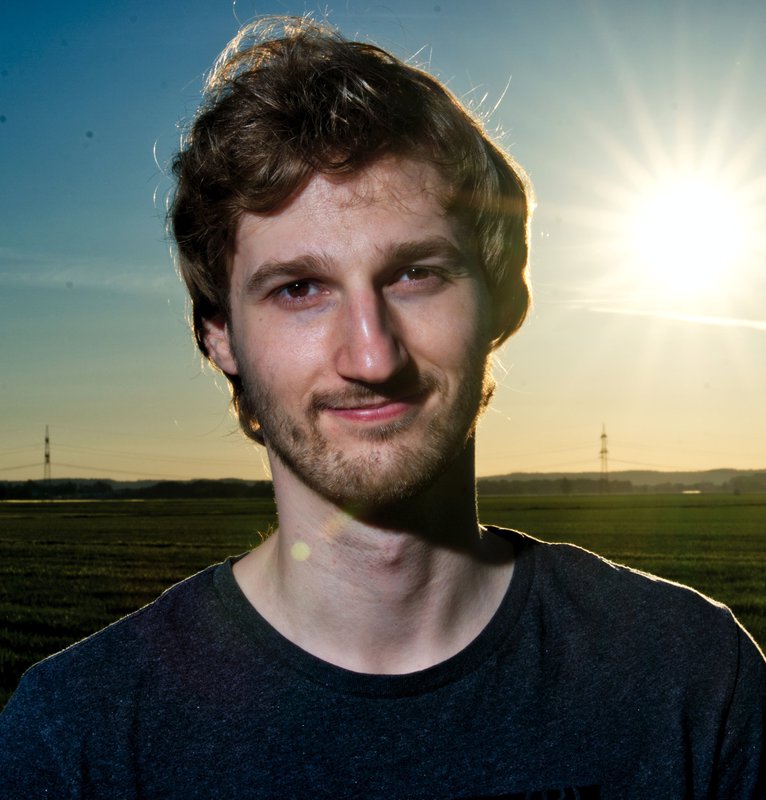}}]{Manuel Milling}
received his Bachelor of Science in Physics and in Computer Science from the University of Augsburg in 2014 and 2015, respectively and his Master of Science in Physics from the same university in 2018.
He is currently a PhD candidate in Computer Science. 
His research interests include machine learning with a particular focus on the development and application of deep learning methodologies.
\end{IEEEbiography}	

\begin{IEEEbiography}[{\includegraphics[width=1in,height=1.25in,clip,keepaspectratio]{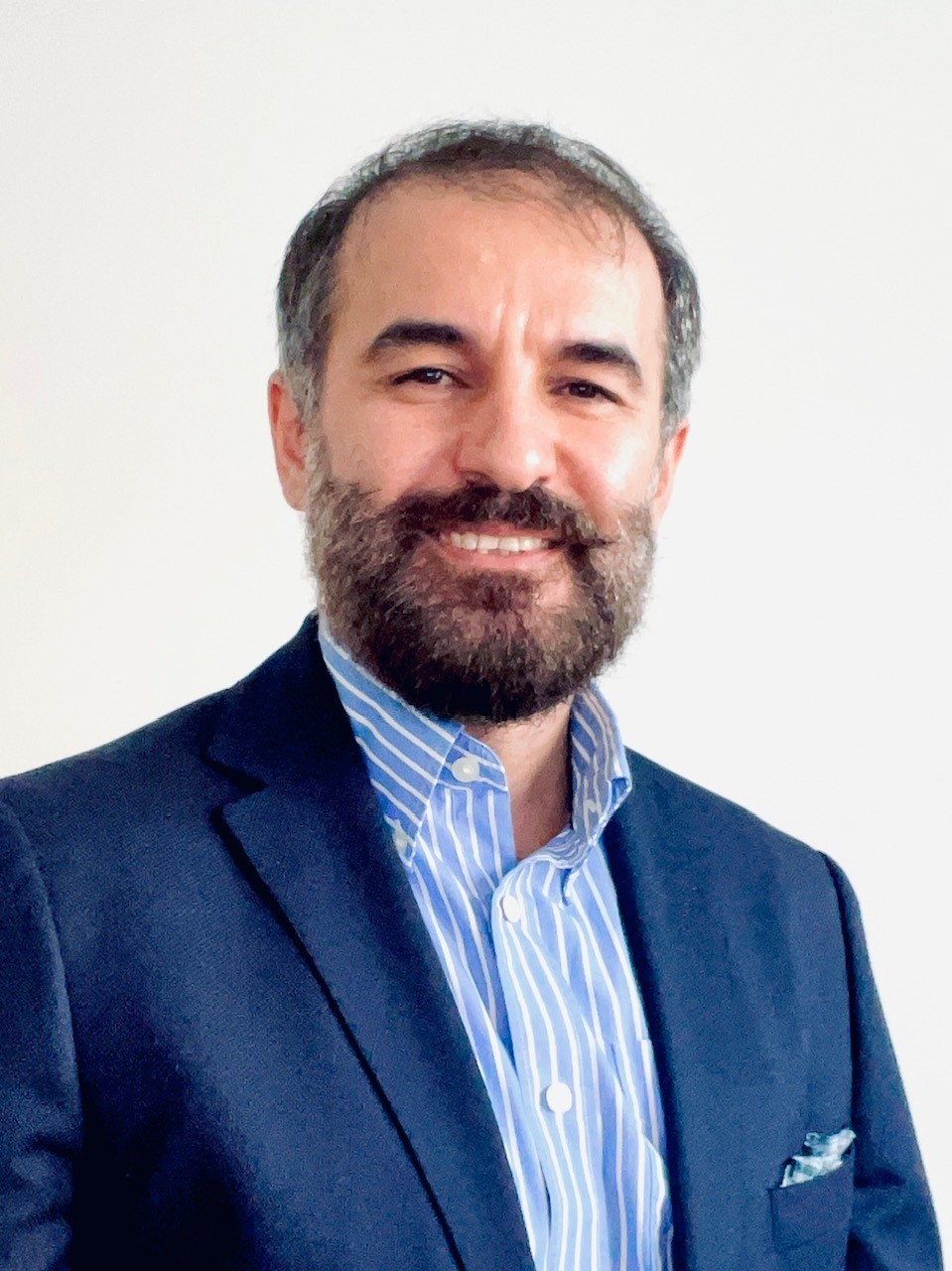}}]{Ilhan Aslan}
 received the Diploma in 2004 from Saarland University in Germany and the Doctoral degree in 2014 at the Center for HCI from Paris-Lodron University Salzburg in Austria. He was an akad.\ Rat ($\sim$assistant professor) at Augsburg University from 2016 onward before joining Huawei Technologies in 2020 as an HCI Expert where he is currently leading an HCI team and managing the Affective Computing \& HCI Innovation Research Lab. His research focus is at the intersection of HCI, IxD, and Affective Computing, exploring the future of human-centred multimedia and multimodal interaction. 
 
\end{IEEEbiography}	

\begin{IEEEbiography}[{\includegraphics[width=1in,height=1.25in,clip,keepaspectratio]{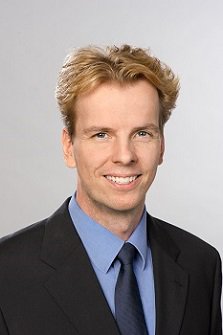}}]{Bj\"orn Schuller}
(M'06, SM'15, Fellow'18) received the Diploma in 1999, the Doctoral degree in 2006, and the Habilitation and Adjunct Teaching Professorship in the subject area of signal processing and machine intelligence in 2012, all in electrical engineering and information technology from Technische Universit\"at M\"unchen (TUM), Munich, Germany. He is Professor of AI and head of GLAM, Imperial College London, London, U.K., and Full Professor and head of the Chair of Embedded Intelligence for Health Care and Wellbeing at the University of Augsburg, Germany. He (co-)authored five books and more than 1\,200 publications in peer reviewed books, journals, and conference proceedings leading to more than 50\,000 citations (h-index = 101). He is Fellow of the AAAC, BCS, ELLIS, IEEE, and ISCA.
\end{IEEEbiography}

\begin{acronym}
\acro{A}[A]{Arousal}
\acro{ABC}[ABC]{Airplane Behaviour Corpus}
\acro{AD}[AD]{Anger Detection}
\acro{AFEW}[AFEW]{Acted Facial Expression in the Wild)}
\acro{AI}[AI]{Artificial Intelligence}
\acro{ANN}[ANN]{Artificial Neural Network}
\acro{ASO}[ASO]{Almost Stochastic Order}
\acro{ASR}[ASR]{Automatic Speech Recognition}

\acro{BN}[BN]{batch normalisation}
\acro{BiLSTM}[BiLSTM]{Bidirectional Long Short-Term Memory}
\acro{BES}[BES]{Burmese Emotional Speech}
\acro{BoAW}[BoAW]{Bag-of-Audio-Words}
\acro{BoDF}[BoDF]{Bag-of-Deep-Feature}
\acro{BoW}[BoW]{Bag-of-Words}

\acro{CASIA}[CASIA]{Speech Emotion Database of the Institute of Automation of the Chinese Academy of Sciences}
\acro{CCC}[CCC]{Concordance Correlation Coefficient}
\acro{CVE}[CVE]{Chinese Vocal Emotions}
\acro{CNN}[CNN]{Con\-vo\-lu\-tion\-al Neural Network}
\acro{CRF}[CRF]{Conditional Random Field}
\acro{CRNN}[CRNN]{Con\-vo\-lu\-tion\-al Recurrent Neural Network}

\acro{DEMoS}[DEMoS]{Database of Elicited Mood in Speech}
\acro{DES}[DES]{Danish Emotional Speech}
\acro{DENS}[DENS]{Dataset for Emotions of Narrative Sequences}
\acro{DNN}[DNN]{Deep Neural Network}
\acro{DS}[DS]{\ds}

\acro{eGeMAPS}[eGeMAPS]{extended version of the Geneva Minimalistic Acoustic Parameter Set}
\acro{EMO-DB}[EMO-DB]{Berlin Database of Emotional Speech}
\acro{EmotiW}[EmotiW 2014]{Emotion in the Wild 2014}
\acro{eNTERFACE}[eNTERFACE]{eNTERFACE'05 Audio-Visual Emotion Database}
\acro{EU-EmoSS}[EU-EmoSS]{EU Emotion Stimulus Set}
\acro{EU-EV}[EU-EV]{EU-Emotion Voice Database}
\acro{EWE}[EWE]{Evaluator-Weighted Estimator}

\acro{AIBO}[FAU Aibo]{FAU Aibo Emotion Corpus}
\acro{FCN}[FCN]{Fully Convolutional Network}
\acro{FFT}[FFT]{fast Fourier transform}

\acro{GAN}[GAN]{Generative Adversarial Network}
\acro{GEMEP}[GEMEP]{Geneva Multimodal Emotion Portrayal}
\acro{GRU}[GRU]{Gated Recurrent Unit}
\acro{GVEESS}[GVEESS]{Geneva Vocal Emotion Expression Stimulus Set}

\acro{IEMOCAP}[IEMOCAP]{Interactive Emotional Dyadic Motion Capture}

\acro{LDA}[LDA]{Latent Dirichlet Allocation}
\acro{LSTM}[LSTM]{Long Short-Term Memory}
\acro{LLD}[LLD]{low-level descriptor}

\acro{MELD}[MELD]{Multimodal EmotionLines Dataset}
\acro{MES}[MES]{Mandarin Emotional Speech}
\acro{MFCC}[MFCC]{Mel-Frequency Cepstral Coefficient}
\acro{MSE}[MSE]{Mean Squared Error}
\acro{MIP}[MIP]{Mood Induction Procedure}
\acro{MLP}[MLP]{Multilayer Perceptron}
\acro{NLP}[NLP]{Natural Language Processing}
\acro{NLU}[NLU]{Natural Language Understanding}
\acro{NMF}[NMF]{Non-negative Matrix Factorization}

\acro{ReLU}[ReLU]{Rectified Linear Unit}
\acro{REMAN}[REMAN]{Relational EMotion ANnotation}
\acro{RMSE}[RMSE]{root mean square error}
\acro{RNN}[RNN]{Recurrent Neural Network}
\acrodefplural{RNN}[RNNs]{Recurrent Neural Networks}

\acro{SER}[SER]{Speech Emotion Recognition}
\acro{SGD}[SGD]{Stochastic Gradient Descent}
\acro{SVM}[SVM]{Support Vector Machine}
\acro{SIMIS}[SIMIS]{Speech in Minimal Invasive Surgery}
\acro{SmartKom}[SmartKom]{SmartKom Multimodal Corpus}
\acro{SEND}[SEND]{Stanford Emotional Narratives Dataset}
\acro{SUSAS}[SUSAS]{Speech Under Simulated and Actual Stress}

\acro{TER}[TER]{Textual Emotion Recognition}
\acro{TTS}[TTS]{Text-to-Speech}

\acro{UAR}[UAR]{Unweighted Average Recall}
\acro{V}[V]{Valence}
\acro{VRNN}[VRNN]{Variational Recurrent Neural Networks}
\acro{WSJ}[WSJ]{Wall Street Journal}
\end{acronym}
\onecolumn

\end{document}